\documentclass[authoryear,preprint,11pt,a4paper]{elsarticle}

\usepackage{amssymb}

\usepackage{array}
\usepackage{svg}
\usepackage{amsmath}
\usepackage{booktabs}
\usepackage{subcaption}
\usepackage{color, soul} 
\usepackage[normalem]{ulem}
\usepackage{float}
\usepackage[colorlinks=true,linkcolor=black, citecolor=blue, urlcolor=blue]{hyperref}
\usepackage[letterpaper,top=2cm,bottom=2cm,left=3cm,right=3cm,marginparwidth=1.75cm]{geometry}
\usepackage{rotating}

\journal{Engineering Applications of Artificial Intelligence}

\begin{document}

\begin{frontmatter}



\title{Artificial Intelligence Approaches for Predictive Maintenance in the Steel Industry: A Survey}


\author[agh,amp]{Jakub Jakubowski}
\author[amp]{Natalia Wojak-Strzelecka}
\author[inesc,uporto_sci]{Rita P. Ribeiro}
\author[rise,hu]{Sepideh Pashami}
\author[uj]{Szymon Bobek}
\author[inesc,uporto_econ]{João Gama}
\author[uj]{Grzegorz J. Nalepa}

\affiliation[agh]{organization={Department of Applied Computer Science, AGH University of Science and Technology},
            addressline={al. Adama Mickiewicza 30}, 
            city={Kraków},
            postcode={30-059}, 
            country={Poland}}

\affiliation[amp]{
    organization={ArcelorMittal Poland},
    addressline={al. Józefa Piłsudskiego 92},
    postcode={41-308},
    city={Dabrowa Gornicza},
    country={Poland}
    }

\affiliation[uj]{organization={Jagiellonian University, Faculty of Physics, Astronomy and Applied Computer Science, Institute of Applied Computer Science, and Jagiellonian Human-Centered AI Lab (JAHCAI), and Mark Kac Center for Complex Systems Research},
            addressline={ul. prof. Stanisława Łojasiewicza 11}, 
            city={Kraków},
            postcode={30-348}, 
            country={Poland}}

\affiliation[rise]{
    organization={RISE Research Institutes of Sweden},
    country={Sweden}}
    
\affiliation[hu]{
    organization={Center for Applied Intelligent Systems Research, Halmstad University},
    city={Halmstad},
    country={Sweden}}

\affiliation[inesc]{
    organization={INESC TEC.},
    postcode={4200-465},
    city={Porto},
    country={Portugal}}

\affiliation[uporto_sci]{
    organization={Faculty of Sciences, University of Porto},
    postcode={4200-465},
    city={Porto},
    country={Portugal}}

\affiliation[uporto_econ]{
    organization={Faculty of Economics, University of Porto},
    postcode={4200-465},
    city={Porto},
    country={Portugal}}

\begin{abstract}
Predictive Maintenance (PdM) emerged as one of the pillars of Industry 4.0, and became crucial for enhancing operational efficiency, allowing to minimize downtime, extend lifespan of equipment, and prevent failures.
A wide range of PdM tasks can be performed using Artificial Intelligence (AI) methods, which often use data generated from industrial sensors.
The steel industry, which is an important branch of the global economy, is one of the potential beneficiaries of this trend, given its large environmental footprint, the globalized nature of the market, and the demanding working conditions.
This survey synthesizes the current state of knowledge in the field of AI-based PdM within the steel industry and is addressed to researchers and practitioners.
We identified 219 articles related to this topic and formulated five research questions, allowing us to gain a global perspective on current trends and the main research gaps.
We examined equipment and facilities subjected to PdM, determined common PdM approaches, and identified trends in the AI methods used to develop these solutions.
We explored the characteristics of the data used in the surveyed articles and assessed the practical implications of the research presented there.
Most of the research focuses on the blast furnace or hot rolling, using data from industrial sensors.
Current trends show increasing interest in the domain, especially in the use of deep learning.
The main challenges include implementing the proposed methods in a production environment, incorporating them into maintenance plans, and enhancing the accessibility and reproducibility of the research.

\end{abstract}




\end{frontmatter}


\newpage
\section{Introduction}
\label{sec:intro}
The practical uses of AI are growing rapidly, and the impact of its applications in many areas is clearly visible.
The industrial landscape has undergone significant transformations in recent decades, driven by integrating digital technologies into manufacturing processes, a phenomenon commonly referred to as Industry 4.0 (I4.0). 
This fourth industrial revolution started a new era of automation, connectivity, and data-driven decision making, enhancing the efficiency and productivity of industrial infrastructures. 
Concurrently, discussions about the industry's future trajectory have resulted in the conceptualization of Industry 5.0 as a potential next phase. 
It emphasizes the synergy between human workers and advanced technologies, advocating a harmonious collaboration between human and artificial intelligence systems~\citep{XU2021530}.
A large amount of data from industry equipment is used to deliver solutions to a specific task. 
An important task in the area of I4.0 is \emph{Predictive Maintenance} (PdM)~\citep{mobley2002introduction}. 
It involves predicting potential equipment failures before they occur, allowing engineers to schedule maintenance activities efficiently. 
Implementing PdM tasks substantially reduces breakdowns, downtimes, and operational costs~\citep{selcuk2017predictive}. 

One of the potential beneficiaries of I4.0 is the steel industry.
Steel, an alloy primarily composed of iron with varying carbon content and different additives, plays a crucial role in the global economy and currently does not have a competitive substitute.
The steel industry is an important supplier of materials for dozens of sectors, including construction, infrastructure, automotive, and manufacturing.
It is also an important employer in many regions, providing millions of jobs around the world.
The distinctive features of the steel industry make it particularly well suited for the implementation of predictive maintenance applications.
First, steel production remains a resource-intensive industry, contributing significantly to air pollution and waste generation, which poses environmental challenges~\citep{met10091117}.
The globalized nature of the steel market makes it susceptible to economic and geopolitical factors. Therefore, optimal utilization of resources and production capacities is necessary to ensure economic stability.
Finally, heavy industry is typically an area of increased risk; therefore, optimal maintenance of equipment can increase the safety of employees.
All these characteristics show the demanding picture of the steel industry, emphasizing the need for intelligent process monitoring.
The integration of predictive maintenance solutions can increase environmental, economic, and personnel safety and help build a modern workplace.

We did not find any previous work synthesizing the current state of knowledge in AI-based PdM in the steel industry.
Despite the absence of directly comparable works, some loosely related studies offer insight into tangential areas.
~\citet{Gajdzik2021} explored the practical problems of steel manufacturers in transitioning to I4.0.
~\citet{Cemernek2022} reviewed the current state of knowledge within the application of ML techniques in continuous casting, which is one of the steel production steps.
Therefore, we believe that our work is novel, positioning it as a promising foundation for researchers aiming to advance PdM methodologies for the steel sector and for practitioners seeking to implement such solutions in their facilities.

Our expertise in the surveyed domain allows us to conduct a thorough and comprehensive survey.
The authors of this article are involved in international research projects within the fields of I4.0 and AI.
One of such projects is the CHIST-ERA XPM project\footnote{See \url{https://xpm.project.uj.edu.pl}.}, which is focused on the development of novel Explainable Artificial Intelligence (XAI) techniques for PdM.
The second example is the CHIST-ERA PACMEL project\footnote{See \url{https://www.chistera.eu/projects/pacmel}.}, which involved the discovery of knowledge from industrial sensors and logs using data mining techniques.
In both projects, we cooperated with ArcelorMittal 
which is one of the most important global corporations in the steel industry.
Furthermore, the leading authors of the survey have been working professionally in this company.
This provides us with valuable insight into the requirements of both research and business, enabling us to approach the field from various perspectives.



The main contributions of this study are the following:
a) Identification of the research papers devoted to AI-driven PdM in the steel industry;
b) Synthesis of the current state of knowledge along different dimensions of the subject, i.e., steel industry, PdM and AI;
c) Overview of the surveyed articles with reference to spatial distribution of research, bibliometric analysis and co-authorship network analysis;
d) Determination of the data characteristics used within the surveyed articles, along with its origin and availability;
e) Evaluation of the practical implications of the proposed methods in manufacturing sites, with a main focus on the implementation of the methods and their estimated business profits;
f) Identification of current trends, research gaps, and potential future directions in the development of intelligent PdM solutions for the steel industry.

The remainder of the paper is organized as follows.
In Sect.~\ref{sec:background} we briefly introduce the topics of PdM, AI, and the steel industry.
In Sect.~\ref{sec:methods} we describe the details of the methodology used for the collection of relevant articles and define the research questions that determine the scope of this survey.
Then in Sect.~\ref{sec:overview} we give an overview of the reviewed articles with respect to their bibliometric data.
The specific results of the study, which are based on the defined research questions, are presented in Sect.~\ref{sec:results}.
The discussion of these findings is provided in Sect.~\ref{sec:discussion}.
The paper is concluded in Sect.~\ref{sec:summary}.
As our study is large, we moved some detailed findings to~\ref{app:details}.

\section{Background}
\label{sec:background}

\subsection{What is PdM and how it is related to smart industry}
\label{sec:pdm}

Effective management of the production process requires maintenance, whose objective is to preserve good condition of all assets throughout their lifetime.
Total maintenance costs in the industry are estimated between 15\% and 70\% of the value of the end product~\citep{Thomas2018maintenance}, and include repairs and replacement of assets, downtime, delays, and labor. 
In addition to financial losses, poor maintenance can lead to environmental or safety hazards. 

In general, there are three main approaches to the maintenance of equipment.
The \emph{reactive}, or \emph{run-to-failure}, approach assumes that no specific actions are taken toward a given asset until it fails or prevents further production.
This approach requires little effort from the maintenance team but can be least cost-effective, leading to unexpected downtimes, and may cause significant losses in the damaged asset.
\emph{Preventive} maintenance involves the routine maintenance of the equipment, which is performed at scheduled intervals without considering the actual condition of an asset.
Arguably, this is the most often applied maintenance policy in today's industrial facilities.
\emph{Predictive} maintenance aims to monitor the equipment online to assess the condition of the equipment in real time and optimize maintenance actions, reducing total operational costs and increasing productivity.

Data-driven PdM is a technique that uses data analysis tools and machine learning (ML) algorithms to predict when equipment is likely to fail so that maintenance can be performed before failure occurs. 
This approach can help reduce downtime, increase the useful life of equipment, and improve safety. PdM is a key component of the smart factory concept, which aims to use data and automation to optimize 
industrial 
processes and improve its efficiency.
There are two main groups of PdM tasks according to the time frame they focus on~\citep{pashami_coRR_2023}. The first group deals with diagnostics, which focuses on identifying failures that have already or are currently occurring. The second group deals with prognostics, which aim to predict potential failures that may occur in the future.

Real-time diagnostic tasks require the continuous acquisition and analysis of operating data along with the external variables required to maintain system health and efficiency. 
It also improves the identification of faults and failures. 
A fault could be caused by a temporary situation or an external factor that the equipment can recover from. 
However, a failure can result from multiple consecutive faults, such as the equipment not recovering or disrupting its operation.
In the context of diagnostics, there are different PdM tasks that progress from low-level data layers to more high-level conceptual ones~\citep{Chandola2009anomaly},\citep{pashami_coRR_2023},\citep{Chemweno2016rca}:
\emph{Anomaly Detection}~\footnote{In this article term \emph{anomaly detection} refers to PdM task, to describe ML task we use term \emph{outlier detection}.} detects unexpected deviations from the normal process behaviour, often measured using anomaly score, which determines the predicted degree of anomaly.
\emph{Fault Detection} identifies if the system is experiencing faults. Generally, it can be considered an anomaly that goes beyond the desired operating conditions.
\emph{Fault Diagnosis} pinpoints the specific fault or malfunction by analyzing the data patterns and inferring the root causes.
\emph{Fault Identification} goes beyond diagnosis to precisely identify the type of fault and quantify its severity.
\emph{Fault Isolation} identifies the specific component or subsystem that causes the identified fault.
\emph{Root Cause Analysis (RCA)} identifies the focal root causes of equipment failures.
It can be considered as the most in-depth analysis of the fault, as it aims to determine the underlying causes of the problems.
%
An integrated approach between these tasks is of utmost importance for an effective maintenance approach. 
Early detection, identification, and resolution of faults can lead to significant cost savings and efficiency improvements.

\emph{Process} and \emph{condition monitoring} are concepts related to diagnostics whose objective is to control the state of the equipment or production line.
Although these approaches are similar to anomaly detection or fault detection, they stand as separate concepts in the literature.
Condition monitoring focuses more on the individual asset or subsystem~\citep{Wakiru2019cm}, and process monitoring usually reaches a wider scope~\citep{Severson2015pm}.

PdM system can also aim to monitor the quality of the product itself, which is often called \emph{defect detection}.
This approach helps maintain good properties of the manufactured product and can indicate issues with the process, resulting in an increase in the number of defects.  
However, defects in the product are not necessarily connected with equipment failure but can be the effect of invalid processing or low-quality input material.

From a prognostics perspective, complementary tasks include:
\emph{Health Indicator} (HI) assessing the current state of a system or machine, often using condition monitoring data and event data. Based on the estimated level of degradation, a prognosis of the future state can be made~\citep{pashami_coRR_2023}.
\emph{Remaining Useful Life} (RUL) estimating the time for a machine or system to function before it requires maintenance or replacement~\citep{Okoh2014rul}.

These concepts work together to provide a comprehensive understanding of the health and useful life of a system. 
The estimation of HI provides real-time information on the current state of the system. 
RUL estimates the remaining life of the system before maintenance is required. 
On the basis of these estimates, failure prediction can be attained by setting thresholds such as end-of-life or time-to-failure.

Ultimately, the approaches described above can be integrated to enable effective predictive maintenance and decision making in industrial machinery.
Information on the current and future state of equipment can be used in \emph{scheduling} tasks to optimize production and maintenance plans~\citep{Zhai2021scheduling}.

\subsection{Commonly used AI methods for PdM} 
\label{sec:aimet}
The predictive capabilities of AI have provided a successful alternative to classical rule-based maintenance. 
The use of AI methods, and especially ML, for predictive maintenance in industrial applications was analyzed in many survey articles~\citep{Surucu2023_PdM_Survey,Zhang2019_PdM_Survey,Zonta2020_PdM_Survey}.
The works mentioned are mainly focused on utilizing different ML techniques to perform PdM tasks, emphasizing the importance of this AI subfield.
\citet{Zonta2020_PdM_Survey} also consider physical model-based and knowledge-based approaches along with data-driven methods (which encompass ML) in their study, however, they claim that the latter group constitutes the majority of research.

Within this survey, we classified the ML algorithms into several groups: 
\begin{itemize}
    \item \emph{Neural Networks} (NN) are brain-inspired algorithms in which thousands, or even billions, of nodes form interconnected layers, transforming the input data into the desired output.
    Neural networks, especially deep learning approaches, are arguably the most versatile and have the broadest application area of all AI methods. Within this group, we can find relatively simple architectures, e.g. Multi-layer Perceptron (MLP) networks, as well as more advanced ones, including Recurrent Neural Networks (RNN), Convolutional Neural Networks (CNN), or Transformer.
    \item \emph{Support Vector Machines} (SVM) are models that find a hyperplane that separates different classes of observations in a high-dimensional space.
    \item \emph{Tree-based} models recursively partition a feature space into subsets based on the most informative features (or using random cuts in special cases).
    A boosting method is often used, in which the predictions of many weak learners are converted into one strong prediction, to increase the predictive capabilities and generalization.
    \item \emph{Probabilistic models} capture uncertainty by modelling data distributions or sequences using probabilistic frameworks, commonly applied in areas such as anomaly detection.
    \item \emph{Clustering} algorithms group similar data points according to their characteristics, allowing data exploration and pattern discovery in unlabeled data sets.
\end{itemize}

Some of the data-driven approaches, which are not within the family of ML but can serve as an alternative, include \emph{statistical} methods, \emph{signal processing}, and \emph{dimensionality reduction}.

Besides the algorithm used, ML methods can be categorized, based on the characteristics of the training procedure, into supervised, unsupervised, or semi-supervised.
\emph{Supervised} approaches require a target label for each observation used for training. 
The algorithm minimizes the discrepancy between the provided label and the prediction of the model.
Popular examples of supervised learning are classification and regression.
On the other hand, \emph{unsupervised} methods learn the patterns in the data provided without a target, which makes them particularly useful in the case of unlabeled data.
In the context of PdM, unsupervised learning can be applied successfully to tasks such as outlier detection or clustering.
Hybrid approaches, which do not fall under any of the above definitions, are referred to as \emph{semi-supervised} learning.
Most ML algorithms are generally not strictly linked to any specific learning approach.

\emph{Regression} is a type of supervised learning that creates a mapping of the desired variables.
One of the PdM tasks often expressed as regression is estimating the RUL of the equipment, which is an essential goal in PdM. 
Predicting RUL requires labeled data that associate time with failure for each measurement. 
However, labeling measurements recorded after the most recent failure is challenging due to the absence of prior knowledge about the timing of the subsequent failure \citep{mashhadi_AS_2020}. 
As a result, certain assumptions are needed to approximate the regression function using ML techniques, e.g. Recurrent Neural Networks \citep{mashhadi_AS_2020}.
Continuing with supervised learning, \emph{Classification} techniques are the go-to approach to determine whether a failure will occur within a specified prediction horizon \citep{PRYTZ2015139}. 
Random forest \citep{PRYTZ2015139} performs well, mainly when failures are infrequent. 
It is important to note that a variety of techniques, such as neural networks and support vector machines \citep{CARVALHO2019106024,Han_2024,Ince_2016}, can also be used similarly for fault detection. 

Various unsupervised \emph{outlier detection} methods, such as one-class support vector machines (OCSVM)~\citep{Scholkopf1999_ocsvm} and Isolation Forest \citep{Liu2008iforest}, are used to identify abnormal patterns or events within the data. 
These methods can detect anomalies in the sense of unexpected patterns or events that may indicate potential failures by comparing them with normal data. 
This process assigns anomaly scores to data points, and if the score exceeds a threshold, an alert is triggered. 
These methods have often been used in scenarios where the labeled data (e.g., failure instances of equipment) are rare or where the various types of potential failure are not fully understood. 
Semi-supervised autoencoder-based (AE) anomaly detection has recently gained attention \citep{Hojjati_2023,serradilla2021adaptable}, particularly for leveraging unlabeled data in fault diagnostics.
These approaches aim to create an encoder and decoder neural network to reproduce the data. The idea is that anomalies should exhibit higher reconstruction errors of the autoencoder compared to normal data to indicate abnormal behavior. 

Within the ML, there are several advanced concepts that can support the development of PdM methods.
Techniques such as \emph{transfer learning} or \emph{domain adaptation}~\citep{TAGHIYARRENANI2023119907}, aim to minimize the need to label data while simultaneously capturing the diversity in the data distribution. 
Various Neural Network architectures are frequently used in this context~\citep{Ganin_JMLR_2016}. 
When solving real-world problems, a model developed for a system can be utilized to adapt to a new system or condition, where data is often more scarce~\citep{TAGHIYARRENANI2023119907}. 
\emph{Reinforcement learning} is a concept in which an intelligent agent learns to make decisions by interacting with the environment and receiving feedback in the form of penalties (in the case of wrong decisions) or rewards (in the case of good decisions). 
It is another way to handle a small amount of labeled data~\citep{QIAN2022104401,YAN2023364} by interactively enhancing the learning of the labeled data while sampling unlabeled data simultaneously. 

\emph{Explainable Artificial Intelligence} (XAI) is a notable advancement in ML techniques designed to describe the decision of black-box models, such as neural networks. 
The explainability aspect is essential to create trust in ML results, particularly in critical domains such as PdM, where AI-based decision support systems must earn the trust of human experts~\citep{pashami_coRR_2023}. 
Feature importance explanation methods such as Shapley values~\citep{Alabdallah2022,Hasan2021,lundberg2017} are a way to indicate variables that cause a failure state.

In addition to ML techniques, several other AI approaches can prove useful for PdM.
AI-based \emph{optimization} methods, e.g. Genetic Algorithms (GA) or Particle Swarm Optimization (PSO), try to find the best possible solution from the subset of all feasible solutions.
In PdM, these methods can be utilized to solve scheduling problems, e.g. optimizing maintenance strategy.
In addition, they can be used to optimize the architectures of ML methods, improving their performance.
\emph{Knowledge-based} methods, e.g. expert systems or fuzzy logic, use domain expertise and historical data to develop inference mechanisms.

\subsection{Steel production}
\label{sec:steel_production}
\begin{figure*}[ht]
    \centering
    \includegraphics[width=0.9\textwidth]{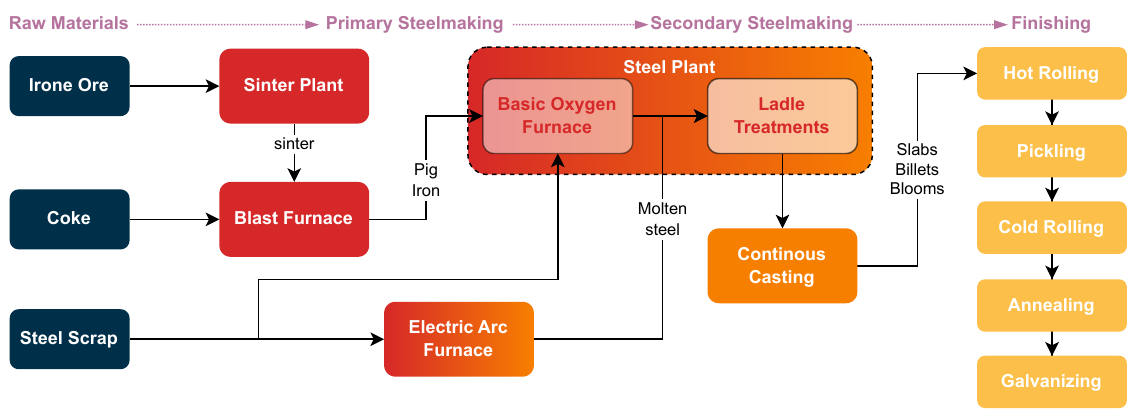}
    \caption{Steel production steps}
    \label{fig:steel_steps}
\end{figure*}

The production of steel is a complex process that involves several manufacturing facilities.
The three main materials used to produce steel are \emph{iron ore}, \emph{coke}, and \emph{steel scrap}.
The extraction and refinement of raw materials (iron ore mining, coking, etc.), which are preliminary production steps, are not within the scope of this survey; therefore, we do not discuss them.
Fig.~\ref{fig:steel_steps} visually represents the typical stages of the steel manufacturing process. 

The first stage of production considered in our study is the sintering of iron ore in a \emph{sinter plant}.
The objective of this process is to agglomerate iron ore particles into a porous mass, by mixing it with other materials, such as limestone and coke breeze, and ignite them in a sintering furnace.
The resulting material is an optimal input for the next steelmaking stage.

\begin{figure}
    \centering
    \includegraphics[width=0.7\columnwidth]{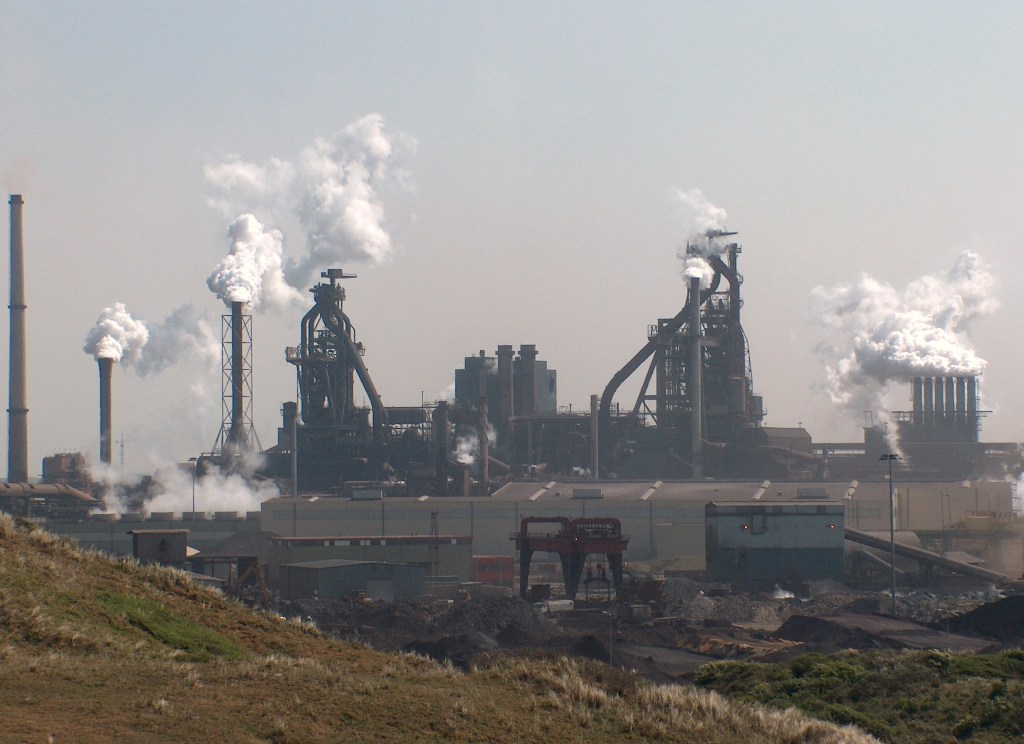}
    \caption{View on the integrated steel factory in Netherlands~\citep{steel_factory_image}}
    \label{fig:img_steel_factory}
\end{figure}

Sinter and coke serve as fundamental inputs supplied to \emph{Blast Furnace} (BF), where they undergo transformation into pig iron.
A BF is a large cylindrical furnace that operates by continuously feeding raw materials to the top of the furnace, where they undergo intense heat and chemical reactions, including smelting, resulting in the extraction of molten iron and the formation of slag, which is tapped off separately.
Fig.~\ref{fig:img_steel_factory} presents a view of an exemplary integrated steel factory, where the two highest towers are BFs.

As an alternative to BF processes, molten steel can be obtained in \emph{Electric Arc Furnace} (EAF), which melts the steel scrap using high-power electric arcs.
It does not require the input of raw materials, and the energy can be supplied from renewable sources, making this technology generally more environmentally friendly compared to the BF process.

The next production steps are performed within \emph{steel plant}.
To produce steel from pig iron, a \emph{Basic Oxygen Furnace} (BOF) is used, in which pure oxygen is blown into a vessel with pig iron.
Following primary steelmaking, which ends with the production of crude steel, various refining processes are performed to adjust the composition and remove impurities.
They are referred to as secondary metallurgy and include e.g. ladle metallurgy, vacuum degassing, or argon stirring. 
Throughout these processes, steel is continuously kept in molten form.

To convert steel into semi-finished goods (SFG), for example, slabs, billets or blooms, \emph{Continuous Casting Machine} (CCM) is used.
Molten steel flows from the ladle through the tundish into a water-cooled mold where it solidifies and is continuously withdrawn, enabling production at a constant speed. 
At the end of the Continuous Casting Machine (CCM), a solid steel product is cut using a torch cutter, resulting in the production of SFG.

Following the casting process, the SFG is transferred to \emph{Hot Rolling Mill} (HRM), where it undergoes further processing.
In HRM, SFG is first reheated in a furnace, reshaped in rolling stands, and cooled down.
Rolling reduces the thickness and width of the material, resulting in its elongation.
In the case of certain types of steel products, e.g. strips or wires, the products are coiled after hot rolling.
In some cases hot-rolling is the final step of steel manufacturing process, but steel may also undergo several further finishing processes to e.g. give the required shape, improve mechanical properties, or increase anti-corrosion resistance.

Typically, after hot rolling, steel is processed at \emph{Pickling Line}, which involves immersion in a bath of acidic solution to remove impurities from the hot rolling process.
Subsequently, a thin layer of oil is applied to the surface to prevent rusting.
If a further reduction in thickness is desired, steel can be rolled in \emph{Cold Rolling Mill} (CRM).
This process is similar to hot rolling, but does not involve preheating of the product, allowing for more precise control of the dimensions.
Due to the internal stresses present in the steel after cold rolling, the \emph{annealing} process is used to restore the steel to its original condition and improve its ductility.
Finally, steel is often subjected to the galvanizing process, which involves coating its surface with a thin layer of zinc. 
This results in increased corrosion protection, which extends the useful life of the product by protecting steel from environmental factors.

The above description of the steel manufacturing process presents the intricate nature of the steel industry.
Proper maintenance of equipment in all facilities is crucial to ensure the quality and continuity of production.
Unplanned downtime at some stage of the production pipeline may influence subsequent production phases.
For example, the unplanned stoppage on the Pickling Line can impact the feasibility of sustaining production on the CRM, annealing, and galvanizing lines, which rely on the uninterrupted supply of input steel.
Therefore, there are many practical and business motivations to develop PdM solutions in steel factories.

\section{Methodology used in the survey}
\label{sec:methods}
This article follows the \emph{Preferred Reporting Items for Systematic Reviews} (PRISMA)~\citep{Page2021_PRISMA} guidelines for the collection of relevant studies to ensure a systematic, comprehensive, and transparent approach to conducting surveys.
The PRISMA framework consists of a checklist and a flow chart for systematic review of articles and is a widely recognized tool for improving the methodological quality of the survey.
We also follow the Search, Appraisal, Synthesis, and Analysis (SALSA) framework, as outlined by \citet{Grant2009_SALSA}, which provides a structured methodology to synthesize and analyze the existing literature.

\subsection{Research questions}
\label{sec:rqs}
The primary objective of this survey is to examine the current state of knowledge on the applications of AI methods to perform PdM tasks within the steel industry.
Given the multidisciplinary nature of this research, we formulate a set of diverse research questions (RQ) that provide different perspectives on the subject matter.
\begin{itemize}
    \item RQ1: What types of installation and equipment are analyzed?
    \item RQ2: What AI methods are used?
    \item RQ3: What types of PdM approaches are used?
    \item RQ4: What are the characteristics of the data used?
    \item RQ5: What is the business impact of the proposed methods?
\end{itemize}
The listed questions serve as the foundation of our study, through which we analyze different dimensions of the subject domain.

The objective of RQ1 is to provide information on the specific areas within the steel industry that received attention from researchers. 
In addressing this question, we aim to clarify details such as the specific installations under investigation and the scope of application (whether it entails an entire installation, specific subsystems, or individual devices).
In addition, we seek to determine whether the methods employed are directed at existing or new installations, along with identifying their geographical location.

RQ2 delves into the characteristics of specific AI methods employed in the examined articles. 
Within this inquiry, our aim is to determine the type of AI methods and their characteristics.
Particularly, in the case of ML methods, we identify the algorithms employed, the type of learning task (e.g. classification, regression, outlier detection), and the selected training approach (e.g. supervised, unsupervised).
Additionally, we seek to validate whether the proposed methodology represents a novel approach, assess the explainability of the method, and determine the authors' efforts toward making the methodology easily reproducible (e.g. by providing the source code).

RQ3 explores the subject from the perspective of PdM tasks.
Within the scope of this inquiry, we aim to obtain insights into the specific type of task addressed, such as anomaly detection, fault diagnosis, RUL estimation, etc.
By formulating this research question, we aim to gain an understanding of the practical implications of the proposed methods for the maintenance of production facilities.

RQ4 aims to identify the characteristics of the data used in the surveyed articles.
The primary focus of this RQ is on categorizing data based on its format, e.g. tabular, time-series, or images. 
In addition, we seek to examine the source of data, i.e., whether the paper uses real-world or synthetic data, and identify if the data were made publicly accessible to other researchers, allowing reproducibility and benchmarking.

The final inquiry, RQ5, seeks to evaluate the practical applications of the proposed research results, %
including the business perspective.
This question involves examining whether the evaluated article involved collaboration with an industrial partner, whether the proposed solution was implemented in a production environment and whether quantitative benefits resulting from its application were presented.

\subsection{Search methodology}
We conducted a systematic review of the literature in four academic databases: Scopus, Web of Science, IEEE Xplore, and ACM Digital Library.
We defined a comprehensive list of keywords that were categorized to align with the thematic framework of this study. 
The first category encompassed terminology associated with AI and covered general concepts such as "machine learning" and specific methods from the domain.
The subsequent keyword category revolved around PdM and its closely related subjects. 
In the final category, we assembled keywords related to the steel industry, encompassing the terminology associated with steelmaking processes and industrial facilities that could be considered.
For the complete list of keywords within these three categories, we refer the readers to Tab.~\ref{tab:keywords}. 

Our objective, when querying the aforementioned databases, was to identify studies that incorporated at least one keyword from each of these categories, ensuring a comprehensive exploration of the research landscape.
We did not provide any restrictions on the publication date of the studies,  allowing for the inclusion of documents dating back to the 1980s and earlier. 
The search was executed in November 2023. 

\begin{table*}
  \centering
  \small
  \begin{tabular}{>{\raggedright}p{0.15\textwidth}|>{\raggedright\arraybackslash}p{0.8\textwidth}}

    \textbf{Category} & \textbf{Keywords} \\
    \hline
    Artificial Intelligence &  machine learning; deep learning; artificial intelligence; neural network; data-driven; data mining; pattern recognition; supervised learning; unsupervised learning; reinforcement learning; clustering; dimensionality reduction; feature extraction; ensemble methods; transfer learning; generative model; time series; data streams; support vector machine; decision tree; random forest; gradient boosting; Bayesian network; generative adversarial network; autoencoder; action mining; survival analysis; decision support; prediction; predictive systems; decision rules; rule-based systems; expert systems; SVM; k-means; k-nearest; XGBoost; XAI; explainability; knowledge discovery; Large Language Models; recommender systems; knowledge base; knowledge-based systems; ontologies; computational ontology; intelligent system; agent system; genetic algorithms; evolutionary algorithms; swarm algorithms; ant colony optimization; heuristic algorithm; heuristic search; DBSCAN; autonomous decision making; RNN; CNN; LSTM; transformer; linear regression; regression; classification; particle swarm optimization; \\
    \hline
    Predictive Maintenance & maintenance; fault detection; fault identification; fault diagnosis; failure prediction; fault prediction; prognostics; diagnostics; condition monitoring; anomaly detection; novelty detection; outlier detection; RUL; remaining useful life; smart monitoring; health management; failure analysis; failure prevention; maintenance optimization; maintenance scheduling; maintenance planning; downtime prevention; root cause analysis; health monitoring; survival analysis; condition assessment; fault isolation; anomaly analysis; outlier analysis; rare events detection; rare events analysis; failure rate estimation; time to failure; reliability analysis; \\
    \hline
    Steel Industry & steel industry; steel manufacturing; steel production; steelmaking; blast furnace; continuous casting; pickling; batch annealing; hot rolling; cold rolling; caster; sinter; electric arc furnace; ladle furnace; continuous annealing; galvanizing; basic oxygen furnace; metallurgy; argon oxygen decarburization; vacuum oxygen decarburization; hot metal; high alloy steel; stainless steel; DRI; vacuum degassing; desulphurization; dephosphorization; steel shop; melt shop; \\
    \hline
  \end{tabular}
  \caption{The complete list of keywords used in the article search procedure}
  \label{tab:keywords}
\end{table*}

\subsection{Selection of relevant studies}
We conducted a multi-step selection process to retrieve previous research articles.
In the first step, we extracted the metadata of all articles that matched the query string.
The metadata included information on the title, year, DOI, authors of the article, as well as its abstract.
We merged all the data for the articles coming from different sources and removed any duplicates.
We also automatically excluded entries that were not a single manuscript, but proceedings from a conference published with a single DOI.

\begin{figure}[ht]
    \centering
    \includegraphics[width=0.5\columnwidth]{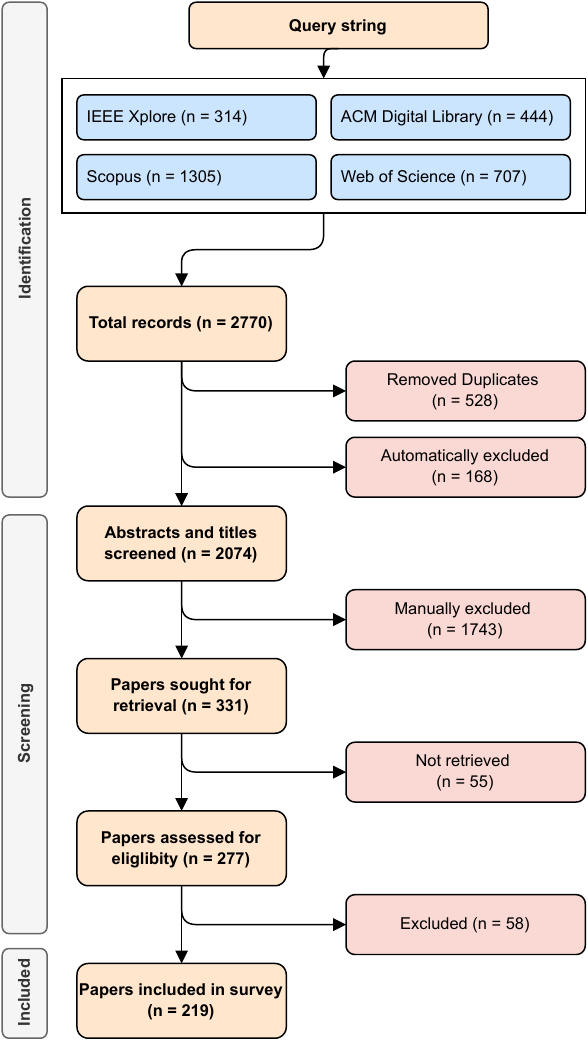}
    \caption{PRISMA workflow presenting the number of retrieved and excluded papers at each stage analysis} 
    \label{fig:prisma_workflow}
\end{figure}

In the next step, two independent reviewers, each possessing expertise in the fields of AI, PdM, and steel industry, conducted a screening process of article titles and abstracts, to assess their relevance to the scope of the survey.
The fundamental inclusion criterion is that the abstract or title must have clear indications of the application of AI in the context of a PdM task within the steel industry.
Each reviewer applied a label that indicated the relevance of the article. 
Articles clearly unrelated to the survey received a label \emph{irrelevant}, while those meeting the inclusion criteria were labeled \emph{relevant}. 
For articles of uncertain relevance, the label \emph{unsure} was applied. 
The assigned labels were compared and the articles marked \emph{relevant} by both reviewers were included in the study.
Subsequently, the studies with an \emph{unsure} label were subjected to individual discussion, and their final labels were assigned by consensus. 
In practice, a significant portion of articles were excluded at this stage due to their lack of relevance to the steel manufacturing industry, instead focusing on other industries or aspects. 

Following the preliminary screening phase, we acquired the complete texts of the selected articles.
During this stage, certain papers were excluded due to retrieval challenges, e.g. lack of a DOI, or being blocked by a paywall from a relatively unrecognized publishers. 
The retrieved papers were subjected to another screening round to ensure eligibility. 
Papers lacking measurable results, short papers, articles mistakenly marked as relevant, or those written in languages other than English were excluded at this point.

A total of 219 articles met the criteria for inclusion in the survey. 
The comprehensive PRISMA workflow diagram, which shows the number of articles retrieved and excluded at each stage of the selection process, is illustrated in Fig.~\ref{fig:prisma_workflow}.

\subsection{Tools}
In the preliminary phase, we explored several tools to improve the literature review process and the extraction of information from scientific papers.
The most robust and valuable tools, based on our assessment, were used in this article and are briefly described below~\footnote{The tools, which we test, but not used in the paper include typeset.io, scite.ai, researchrabbit.ai, scholarcy.}.

The primary analysis of selected papers was performed using a traditional spreadsheet, in which each row represents a specific paper, and each column corresponds to the specific information we aimed to extract (e.g. the name of installation, the AI method, etc.).
The cells within this spreadsheet were completed by two independent reviewers and all discrepancies were discussed to find a common position. 
The representation of all papers and research questions in a table enabled us to summarize the papers in a somewhat manageable manner.
This spreadsheet was later used as input data for Python scripts, which were used to perform synthesis and more advanced data analysis, as well as visualization.

We also performed a co-occurrence analysis using VOSviewer~\citep{vanEck2010}, which allows visualizing textual data in the form of nodes with connections between them and organizing them in clusters, based on the relationships between them.
It can be useful to synthesize the occurrence of keywords or to create a social network of researchers.

Finally, we used the Natural Language Processing (NLP) model~\citep{kulkarni2022learning}, which was trained in the keyword extraction task, to generate keywords for each abstract of a paper.
This was done to provide keywords for all articles and reduce bias with respect to keywords individually supplemented by the individual authors.

We did not use any Generative AI tools. All the analyses were conducted manually based on our expertise and with use of automation procedures we programmed ourselves in Python. 

\section{Overview of Selected Studies}
\label{sec:overview}
In this section, we provide a high-level overview of the studies selected for this survey.
It serves as a quick reference guide, offering readers a snapshot of the key characteristics of the surveyed articles before delving into more detailed analyses in subsequent sections. 
Within this section, we rely on the information stored in the metadata of the articles and use various visualization methods to present different characteristics of the articles.

\subsection{Publication details}
\begin{figure}[ht]
    \centering
    \includegraphics[width=1.0\columnwidth]{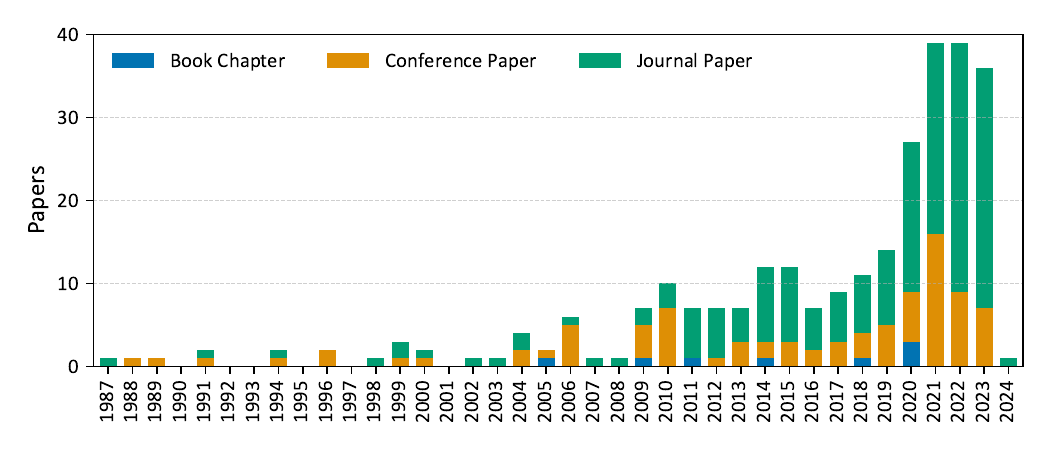}
    \caption{Temporal distribution of analyzed papers based on publication year}
    \label{fig:publications_timeline}
\end{figure}

Fig.~\ref{fig:publications_timeline} illustrates the temporal distribution of the articles surveyed based on their publication years. 
Research in the area dates back to the late 1980s; however, a notable increase in published papers occurred after 2009.
The subsequent substantial increase is evident in recent years, particularly after 2020.
This reflects an increase in the demand for intelligent solutions within the steel industry.
Predominantly, the publications are concentrated in journals, however, a noticeable number of papers is also found in conference proceedings. 
Only a limited number of articles were identified in the book chapters. 

\begin{figure*}[ht]
    \centering
    \begin{subfigure}{0.51\textwidth}
        \includegraphics[width=0.99\columnwidth]{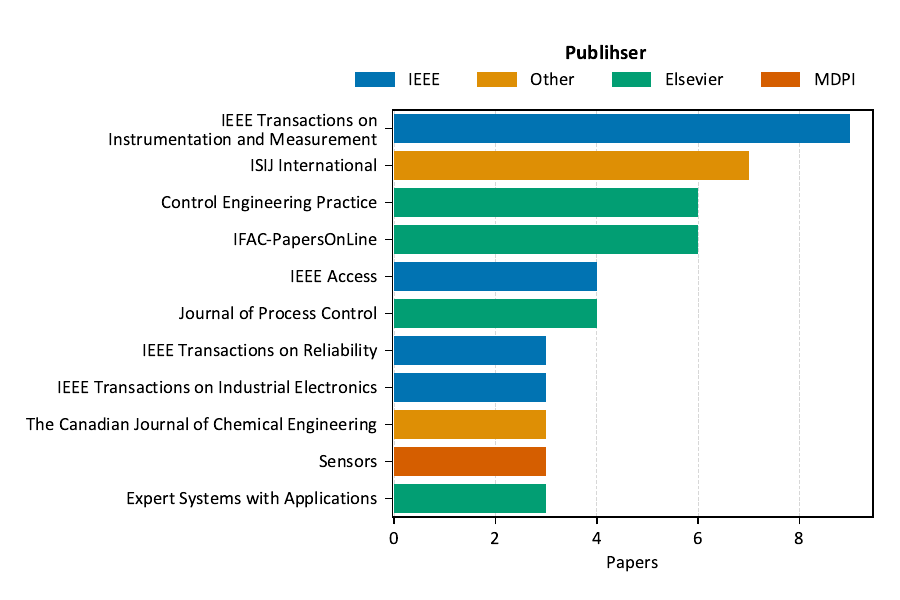}
        \caption{Journals}
        \label{fig:journals_count}
    \end{subfigure}
    \hfill
    \begin{subfigure}{0.48\textwidth}
        \includegraphics[width=0.99\columnwidth]{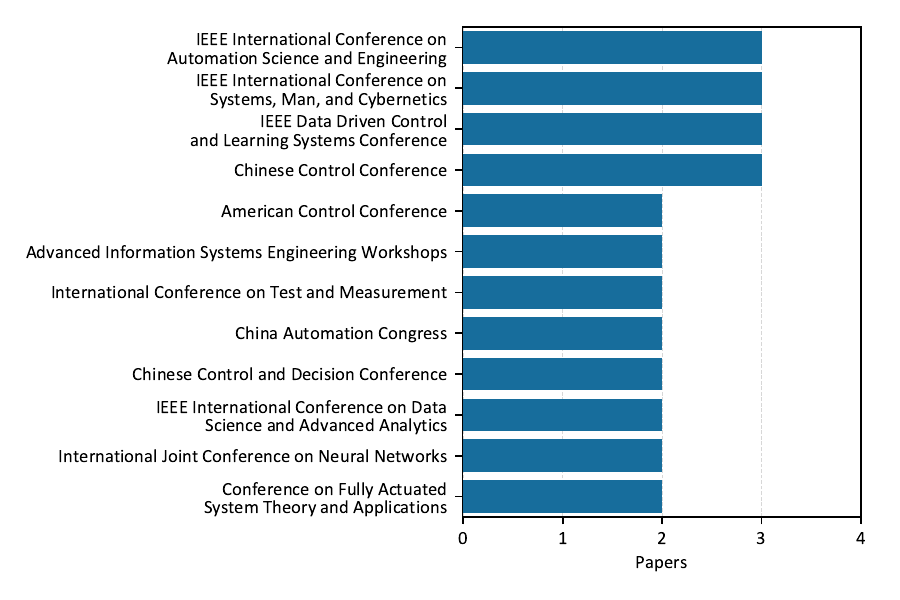}
        \caption{Conferences}
        \label{fig:conferences_count}
    \end{subfigure}
    \caption{Journals and conference, which published the highest number of papers}
    \label{fig:bar_conf_jour}
\end{figure*}

Furthermore, our investigation expanded to an analysis of journals and conferences that exhibited the highest frequency of papers published within the surveyed domain.
Multiple journals emerged as significant contributors, demonstrating a considerable volume of articles. 
Fig.~\ref{fig:journals_count} presents the journals with the highest number of published articles, including those that featured a minimum of three papers included in our survey. 
Noteworthy examples include the "IEEE Transactions on Instrumentation and Measurement" and "ISIJ International". 
The former belongs to the broader group of IEEE journals, while the latter stands as a singular example of a journal aligned with the steel industry. 
A substantial volume of papers is also evident in Elsevier-affiliated journals.
The primary focus of many listed journals revolves around Control Engineering and closely associated fields.
However, there are journals such as "Expert Systems with Applications" or "IEEE Transcations on Industrial Informatics", which are more aligned with the field of Computer Science.
These diverse characteristics of the journals 
highlight the inter-disciplinary nature of the domain.

Similarly, we examined conferences that appear most frequently in the list of surveyed papers.
Fig.~\ref{fig:conferences_count} presents conferences that published a minimum of two papers in their proceedings. 
Although there is no single conference with a notably exceptional number of published articles, a noticeable presence of IEEE-related conferences is observed.
Many of these conferences align with the field of Control Engineering (e.g., "IEEE Data Driven Control and Learning Conference," "Chinese Control Conference"), while instances of conferences related to Computer Science are also apparent (e.g., "International Joint Conference on Neural Networks", "IEEE Conference on Data Science and Advanced Analytics").
In particular, three of the conferences featured in the list are specifically of Chinese origin, suggesting a higher level of research activity within the subject domain from researchers in China.

\begin{figure*}[ht]
    \centering
    \includegraphics[width=1.0\textwidth]{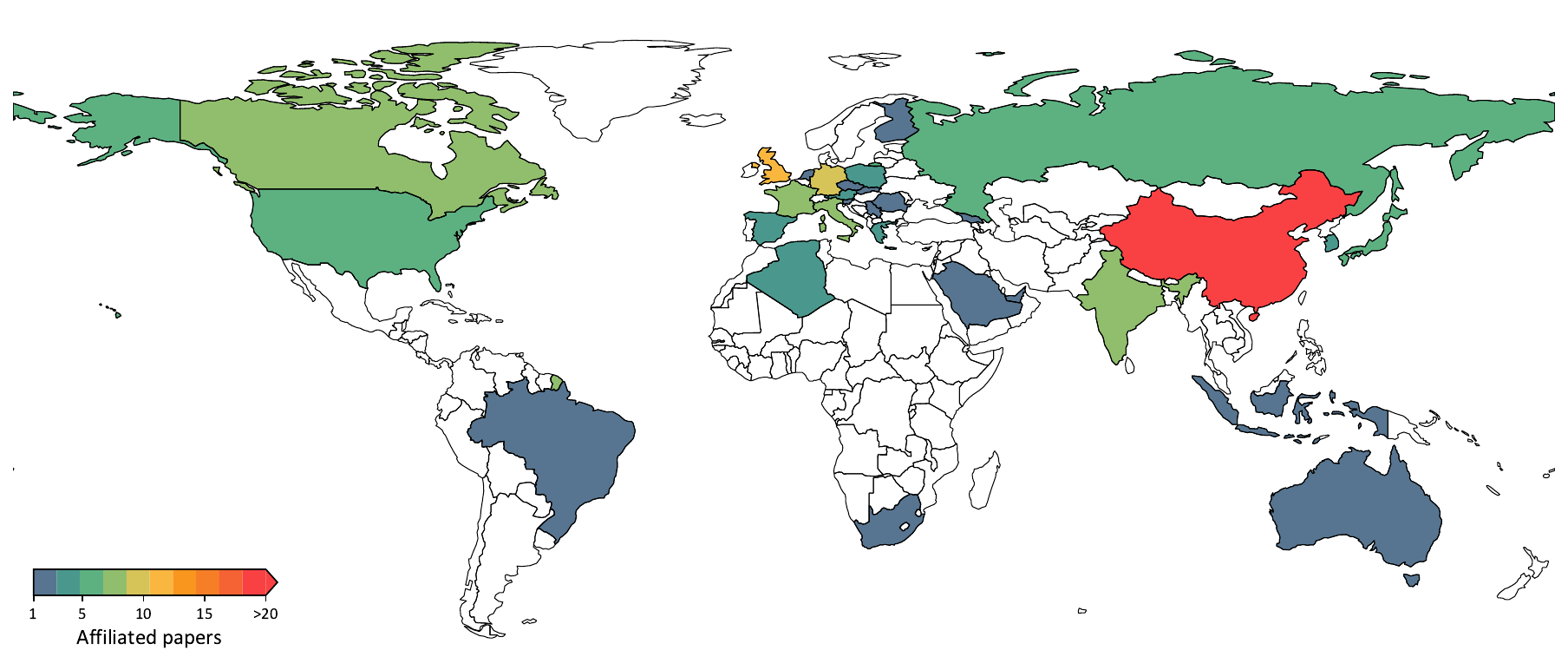}
    \caption{Geographical distribution of research, based on the authors' country of affiliation}
    \label{fig:affiliation_map}
\end{figure*}

The prevalence of research originating in China becomes particularly visible when examining the affiliations of authors. 
The spatial distribution of the affiliated papers between countries is depicted in Fig.~\ref{fig:affiliation_map}. 
We note that more than half of the articles are affiliated with Chinese universities. 
In addition to this, European universities emerge as the second major hub for research activities. 
Noteworthy contributions have also been made from other countries such as Canada, the United States, Russia, India, and Japan.

\subsection{Visualization of keywords and authors association}
\begin{figure}[ht]
    \centering
    \includegraphics[width=1.0\columnwidth]{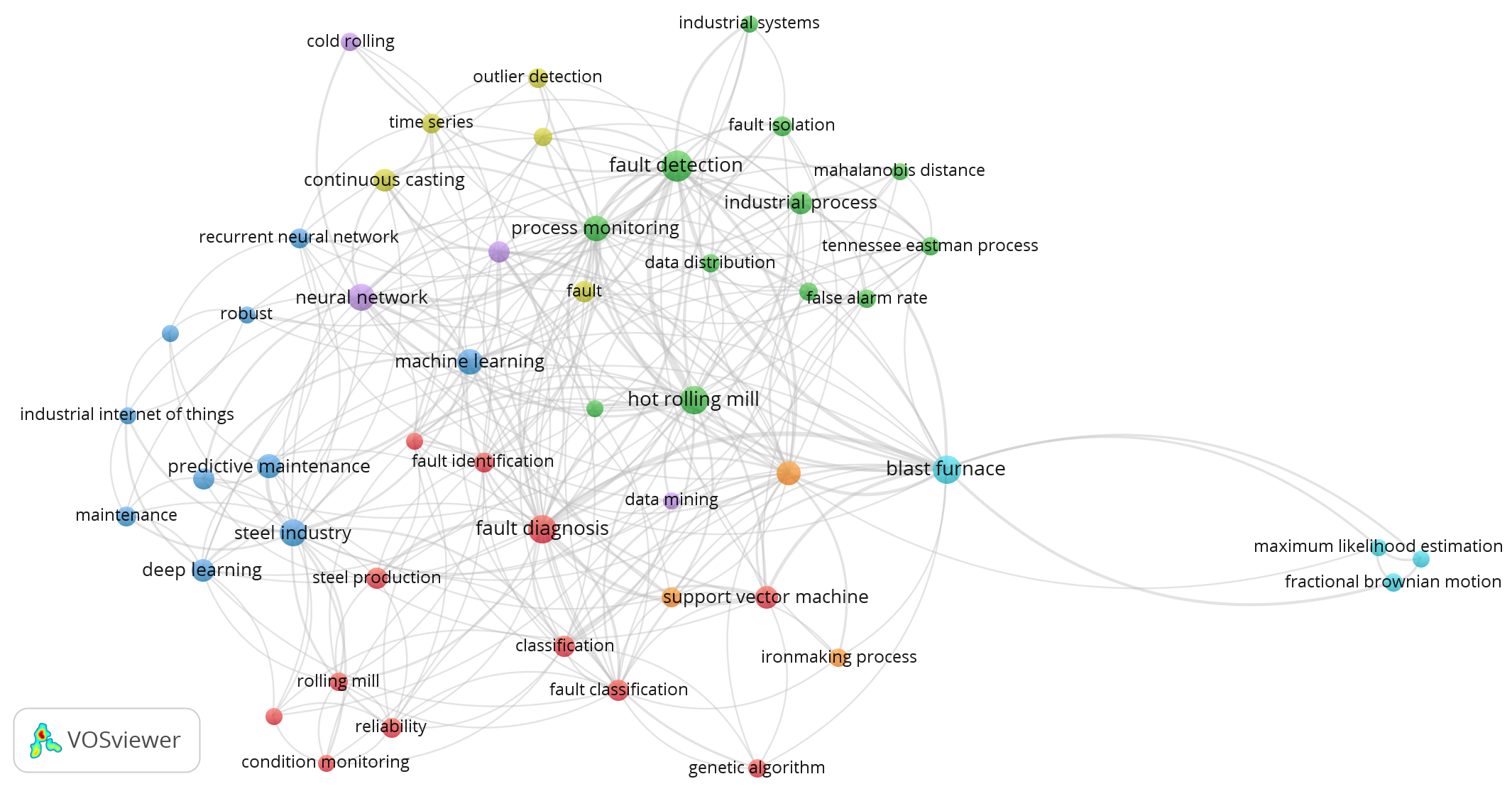}
    \caption{Visual representation of most frequent keywords generated from abstracts and their relationship. Created using VOSViewer}
    \label{fig:vos_keywords}
\end{figure}

To extract information on the main characteristics of the surveyed articles we used NLP model~\citep{kulkarni2022learning} trained on keyword extraction task.
This model was used to generate keywords from the abstracts of all selected articles.
The keywords were then visualized with the use of VOSViewer (Fig.~\ref{fig:vos_keywords}). 
We limited the visualization to keywords to 50 instances, while ensuring that each keyword appeared at least three times.
The colors represent word clusters, and the size of the node indicates the number of keyword occurrences. 

The most commonly used keywords are related to PdM tasks, including terms like \emph{fault detection}, \emph{fault diagnosis}, and \emph{process monitoring}. 
Additionally, \emph{predictive maintenance} is frequently mentioned. 
In the case of AI, terms like \emph{machine learning} and \emph{deep learning} are popular, along with specific algorithms such as \emph{support vector machine,} \emph{neural network,} or \emph{genetic algorithm}. 
In the steel industry context, keywords often pertain to various manufacturing processes and facilities like \emph{blast furnace}, \emph{hot rolling}, and \emph{cold rolling,} as well as broader terms like \emph{steel industry}. 
Despite the focus on steel production, some keywords are more general and encompass manufacturing concepts, like \emph{industrial process} or \emph{industrial internet of things}.
However, we also noted keywords that are not directly relevant to the survey subject or widely recognized approaches. 
For instance, \emph{fractional Brownian motion} refers to a statistical method that can model industrial systems, but it is not commonly associated with PdM applications. 
Similarly, \emph{Tennessee Eastman process} denotes a chemical process described in literature, serving as a benchmark in PdM research, despite not directly aligning with the main focus of the study.

\begin{figure}[ht]
    \centering
    \includegraphics[width=1.0\columnwidth]{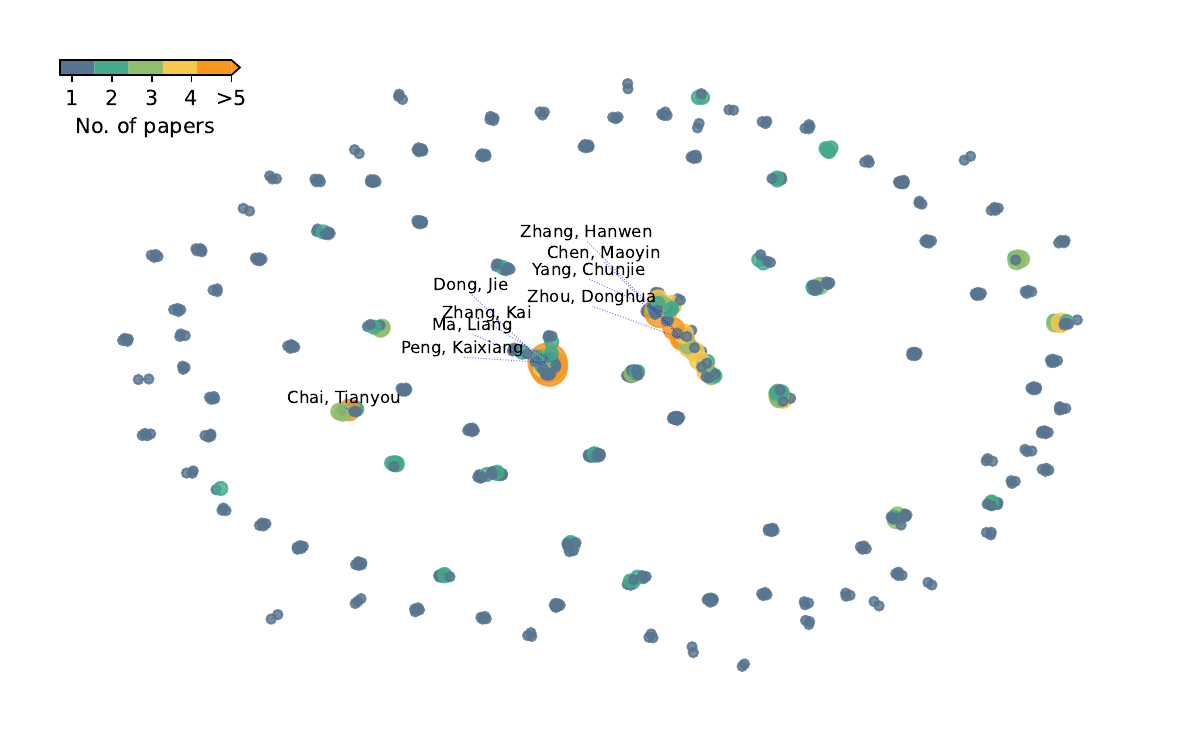}
    \caption{Social Network of authors of analyzed papers. Each node represents a single author, the proximity of two nodes refers to co-authoring of at least one paper. The colors and size of the nodes depicts the number of published papers}
    \label{fig:authors_social}
\end{figure}

Fig.~\ref{fig:authors_social} shows a social network of authors, which presents their impact in the field and the main collaboration teams.
Each node corresponds to an author, and its size and color depend on the number of articles with which this author is affiliated.
The edges between the nodes represent the co-authoring of at least one paper.
The names of the authors who co-authored at least five articles are also included in the graph.
We observe two larger clusters of researchers who worked intensively in the field, and, except for that, the research teams are rather independent and authored only singular papers.
However, several of these teams showed increased research activity, which is reflected in the high number of articles published.

\section{Results of the Study}
\label{sec:results}

This section provides a comprehensive analysis of the articles surveyed.
We present a detailed account of the findings, emphasizing the most common research approaches, as well as more original solutions.
We individually addressed each research question, allowing for a thorough exploration of specific aspects of the study.
Each research question is composed of several more detailed inquiries, which are described in detail in Sect.~\ref{sec:rqs}.
Our aim is to allow the reader to gain a clear and comprehensive understanding of the current state of knowledge and provide a roadmap for navigating within the subject of the study.
The synthesis of the most important findings is presented in Sect.~\ref{sec:discussion}.

\subsection{RQ1: What types of installation and equipment are analyzed?}
\label{sec:plants}

\begin{figure*}[ht]
    \centering
    \includegraphics[width=1.0\textwidth]{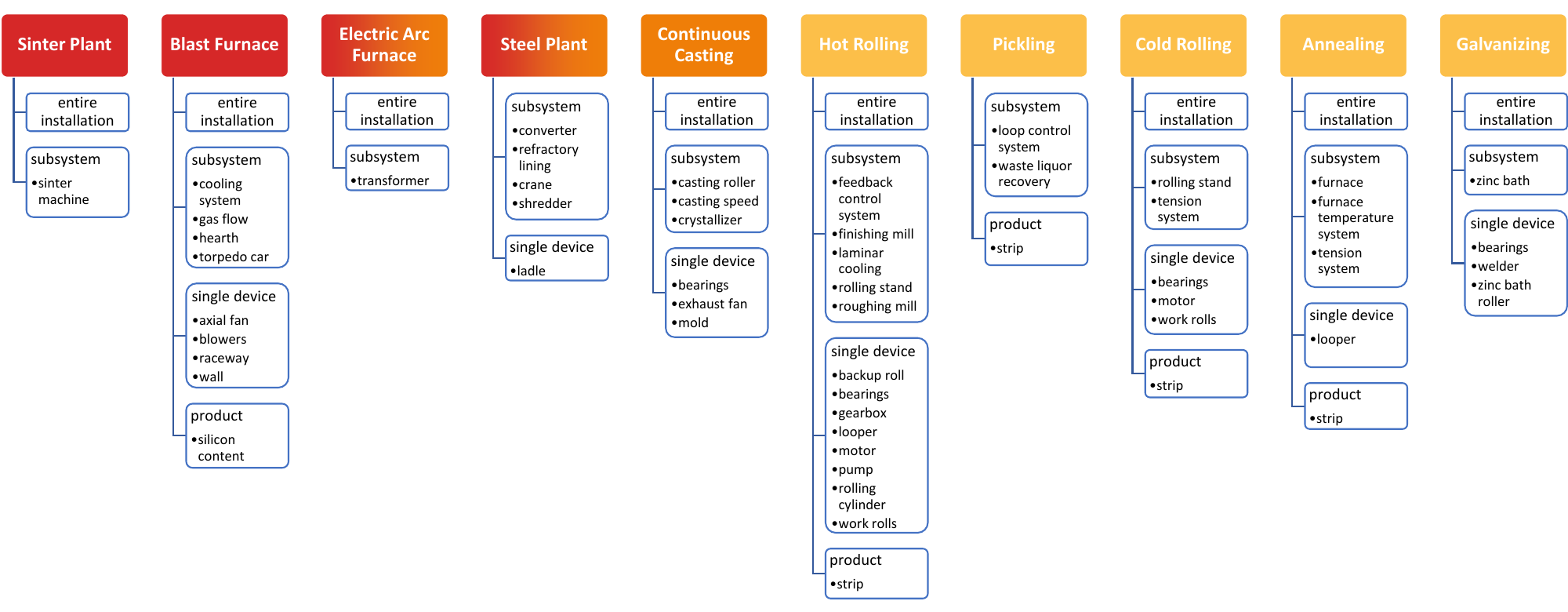}
    \caption{The taxonomy of the installations and equipment subjected to PdM in the analyzed papers}
    \label{fig:installation_taxonomy}
\end{figure*}

To characterize the equipment considered for PdM in steel facilities, we defined a three-level categorization.
At the first level, we extracted information about the type of installation that was the subject of the study.
We then identified the scope of the task, which indicates whether the proposed method was applied to an entire installation, a subsystem, an individual device, or a product.
Finally, we extracted information on the name of the asset that was the subject of the study.
This was not defined for the methods applied to the entire installation, as this would be equal to the name of the installation itself, making this information redundant.
Based on this categorization, we created a taxonomy presented in Fig.~\ref{fig:installation_taxonomy}.

Within this research question, we also extracted information on whether the study is related to existing or new / planned installations and their geographical locations.

\begin{figure}[ht]
    \centering

    \begin{subfigure}{\columnwidth}
        \centering
        \includegraphics[width=0.7\columnwidth]{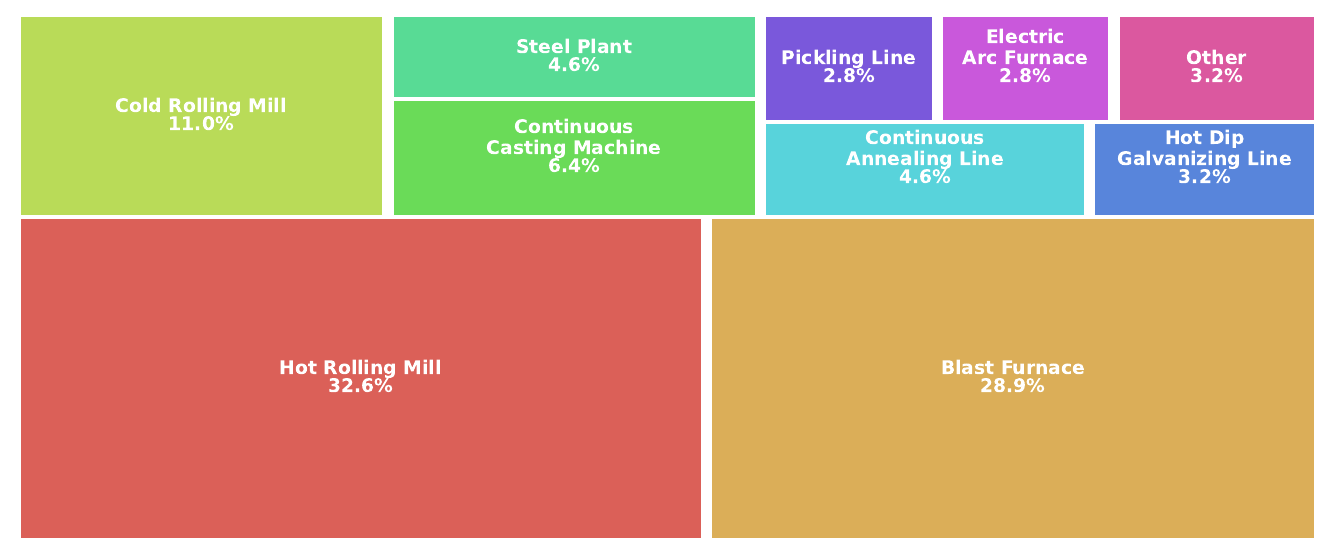}
        \caption{Type of installation}
        \label{fig:treemap_plants}
    \end{subfigure}
    
    \begin{subfigure}{\columnwidth}
        \centering
        \includegraphics[width=0.7\columnwidth]{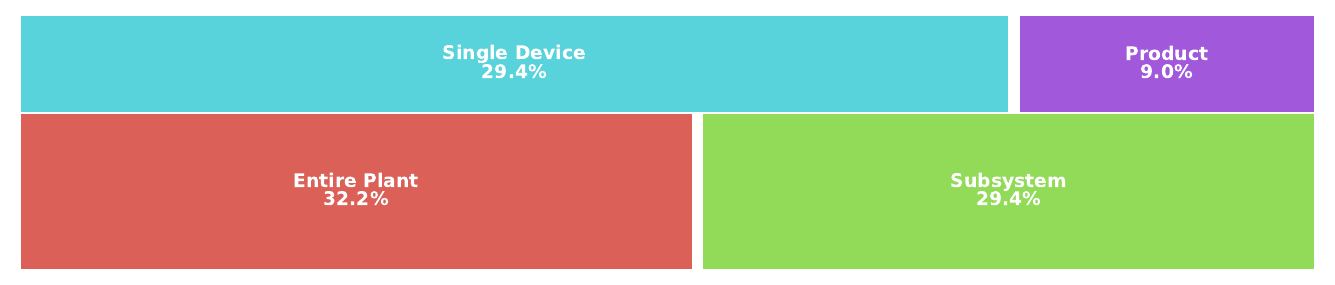}
        \caption{Scope of the task}
        \label{fig:treemap_scope}
    \end{subfigure}
    
    \begin{subfigure}{\columnwidth}
        \centering
        \includegraphics[width=0.7\columnwidth]{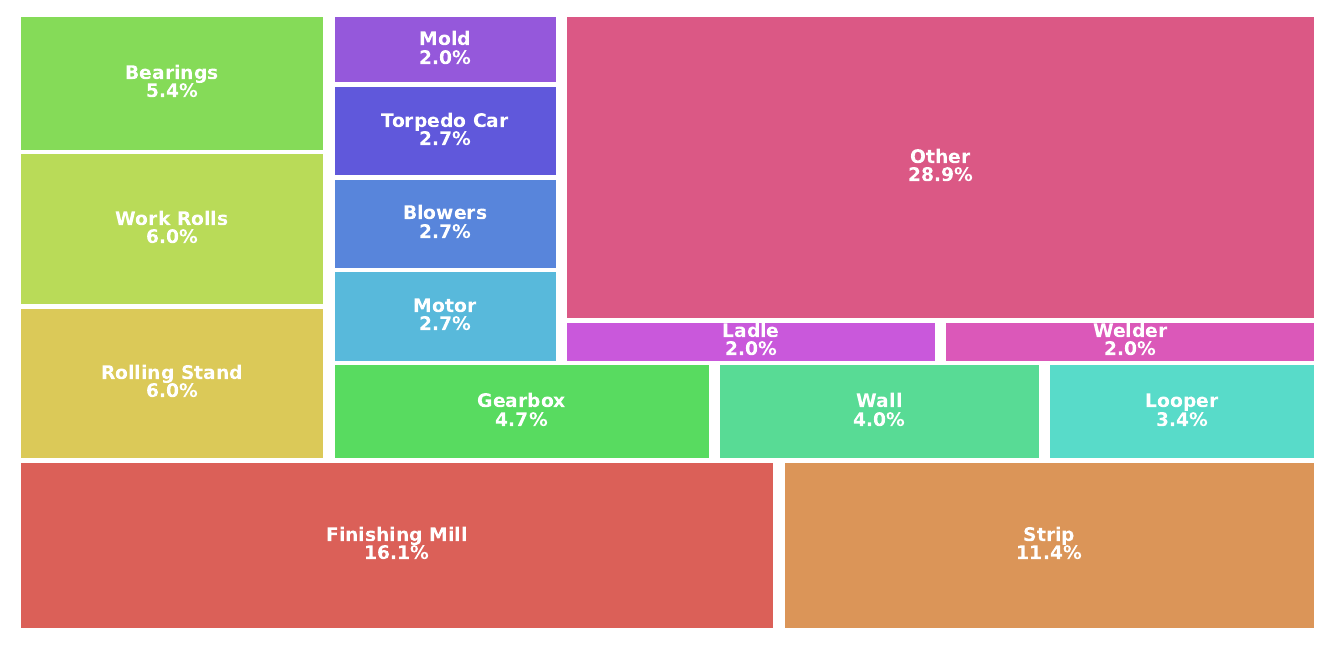}
        \caption{Asset}
        \label{fig:treemap_asset}
    \end{subfigure}

    \caption{Occurrence frequency of areas subjected to PdM with respect to the name of installations, scope of the solution, and name of equipment}
    \label{fig:treemap_rq1}
\end{figure}

\begin{figure}[ht]
    \centering
    \includegraphics[width=0.9\columnwidth]{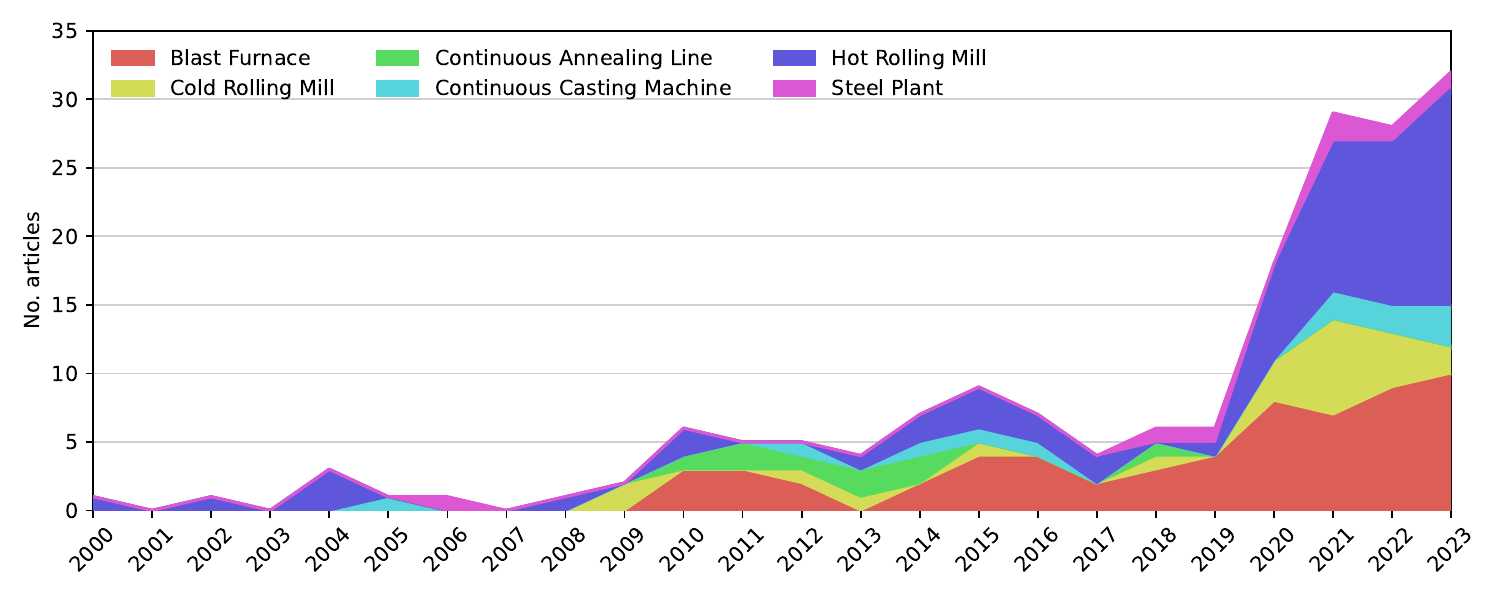}
    \caption{Historical distribution of installations subjected to predictive maintenance}
    \label{fig:plants_timeline}
\end{figure}

Fig.~\ref{fig:treemap_rq1} summarizes the findings in terms of type of installation, scope, and asset.
As observed in Fig.~\ref{fig:treemap_plants} BF and HRM are two major areas of research, accounting for more than 60\% of all research.
Except for these two installations, the CRM is the third most frequent area subjected to PdM, accounting for more than 10\% articles.
A more detailed overview of the installations being studied is presented in Fig.~\ref{fig:plants_timeline}, where their temporal occurrence is visualized.
We observe that until 2020, the highest number of the articles focused on BF.
After this point, the share of papers related to hot rolling increased greatly and outnumbered those devoted to BF. 
An increased interest in cold rolling is also observed.

Moving on to the scope of the studies, which is shown in Fig.~\ref{fig:treemap_scope}, we observe a high diversity in this matter, with each defined category having a significant number of representatives.
However, the smallest number of studies focused solely on the product, while the other categories were approximately evenly represented.

Regarding the occurrence of specific assets, as shown in Fig.~\ref{fig:treemap_asset}, there is a much higher fragmentation of the research, which we find natural considering the variety of equipment used in the steel production process.
The highest amount of research was devoted to \emph{finishing mill}, which is a subsystem of HRM.
Next, significant contributions were made with respect to the steel strip (product).
For individual devices, we observe the presence of typical equipment used in industrial processes, i.e., \emph{bearings}, \emph{gearboxes}, and \emph{motors}, but also devices more specific to steel production, for example \emph{loopers} and \emph{work rolls}.

Our results indicate that most of the research work was dedicated to existing installations.
We identified only three cases in which AI methods were proposed for new facilities.
\citet{marcu_design_2008} designed a fault detection system for a hydraulic looper in HRM, which was validated using data from existing installations, but the intention of the authors is to use this tool for any new or existing facility.
\citet{han_framework_2021} developed a framework for the prediction of RUL in casting rollers based on the digital twin.
\citet{gerz_comparative_2022} proposed a decentralized anomaly detection system, which was tested on the HRM during the commissioning phase.
In some cases, the research was not devoted to the actual manufacturing facility, but the authors performed laboratory experiments, which were the main source of data~\citep{ji_fault_2021,junichi_t_and_yukinori_i_condition_2020,peng_multi-representation_2022,shi_intelligent_2022,spirin_expert_2020,wang_multi-step-ahead_2023,yi_jiangang_and_zeng_peng_analysis_2009,zhao_fault_2021}.

\begin{figure}[ht]
    \centering
    \includegraphics[width=0.9\columnwidth]{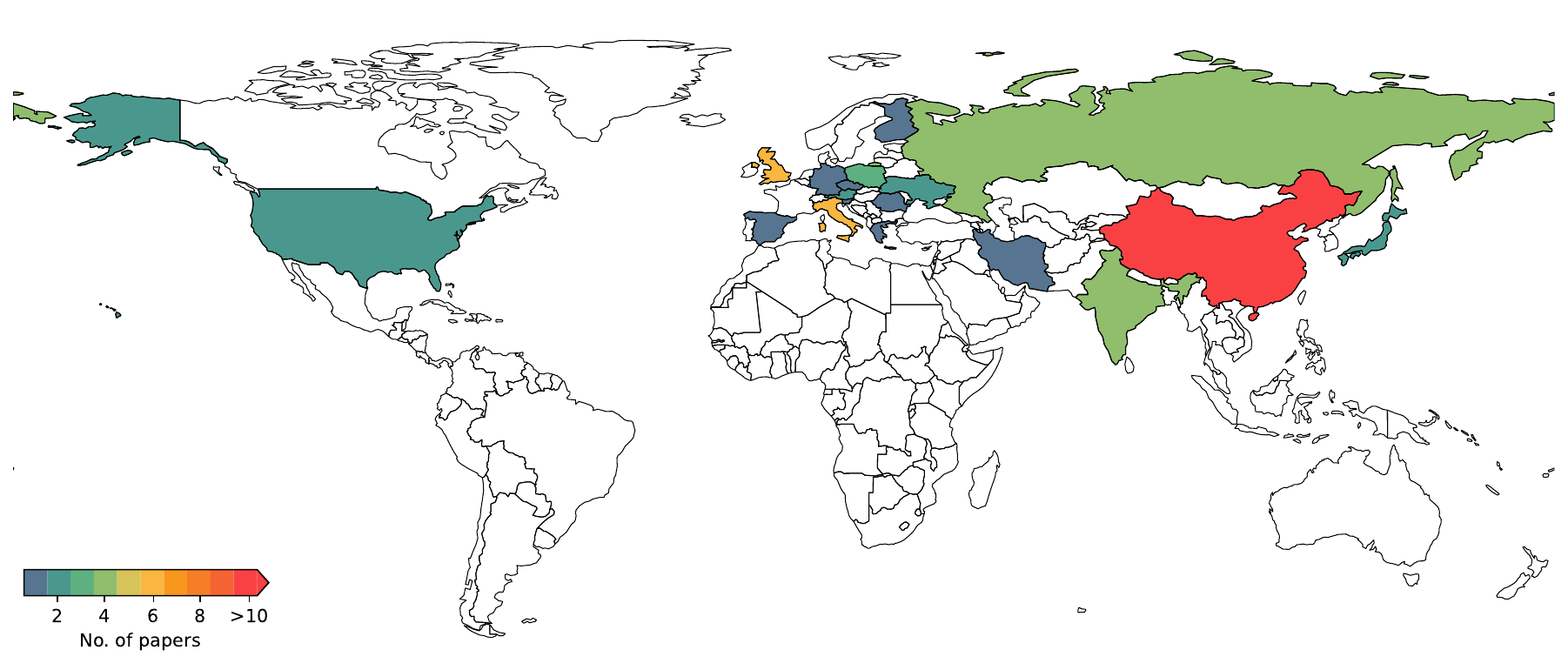}
    \caption{Location of steel facilities used as a case-study in surveyed papers}
    \label{fig:map_location}
\end{figure}

Most studies did not specify the location of the installation, which provided the data.
The count of installations with respect to their geographical location (including only articles that clearly indicated this location) is presented in Fig.~\ref{fig:map_location}.
We find that most of the installations were located in China (more than 30), which is consistent with the affiliation of the authors presented in Fig.~\ref{fig:affiliation_map}.
A significant number of installations subjected to the study were also located in Europe.
Several use cases have been found in Japan, USA, India, and Russia.

To give more detailed information on this manner, in the following subsections we describe the characteristics of the assets subjected to PdM for each installation individually.
Additionally, the complete list of installations with the scope of the equipment considered within them is provided in~\ref{app:details_installations}.

\subsubsection*{Sinter plant}
Sinter plant, which is the first step considered in steel production, did not attract much attention from researchers.
\citet{liu_online_2021} developed a system for sinter batching, which detects abnormal sinter performance.
The primary focus of this paper is not solely anomaly detection, but optimization of the sintering process.
\citet{egorova_diagnostics_2018} applied PCA and ANN to predict faults in the sintering process.

\subsubsection*{Blast Furnace}
Blast Furnace was one of the main areas of research in the articles surveyed.
The specificity of BF, in which the majority of physical and chemical processes take place in one place, that is, the furnace, affects the predominance of research devoted to the development of solutions for the entire installation.
In addition, the harsh environment, along with the complexity and smooth boundaries of the processes that take place inside the furnace, makes it demanding to precisely monitor the conditions inside different parts of the furnace.
Due to the large number of papers on this scope of study, we refer the reader to Tab.~\ref{tab:bf} for a detailed list of articles.

Except for the approach that aims to apply PdM to the entire installation, we identified several studies related to more specific subsystems.
The subsystems that were analyzed the most frequently were torpedo cars, which are used to transport molten steel from BF to steelmaking facilities~\citep{chernyi_application_2022,li_research_2020,wang_research_2021,yemelyanov_application_2021} and furnace walls (see Tab.~\ref{tab:bf}).
Other areas studied include the cooling system~\citep{meneganti_fuzzy_1998}, gas flow subsystem~\citep{zhao_adaptive_2014}, raceway~\citep{puttinger_improving_2019} and the hearth~\citep{wang_shuqiao_and_yuan_yan_and_liang_jingjie_and_zhang_zhuofu_design_2022}.
Regarding individual equipment, blowers~\citep{bo_qualitative_2015,fu_digital_2023,raducan_prediction_2020,sun_research_2012} and axial fan~\citep{zhang_fault_2021} were studied.
Furthermore, we identified two articles that focus on monitoring the silicon content in the final product of the BF process~\citep{liu_optional_2011,weng_online_2023}.

\subsubsection*{Electric Arc Furnace}
Studies related to EAF focused on the ironmaking process in the furnace or transformer, which is one of the most crucial subsystems.
In terms of articles devoted to the furnace, \citet{wang_approach_2015} proposed using wavelet analysis and ML to detect anomalies in the EAF process.
Later they proposed~\citep{wang_detecting_2018} alternative approach for this task ~\citep{wang_detecting_2018}, which utilized clustering methods.
\citet{choi_method_2020} aimed to predict the time to the next maintenance of the EAF based on the sensor data collected.
For the transformer subsystem, \citet{dehghan_marvasti_fault_2014} proposed a method to detect faults on the secondary side of the EAF based on primary side data, while \citet{shiyan_vibration-based_2019} used vibration data to monitor this asset.

\subsubsection*{Steel Plant}
Under the term \textit{Steel Plant}, we included all metallurgical processes that occur after BF, but before casting.
Two studies were related to the converter process, which takes place in the BOF; Chistyakova and Novozhilova~\citep{chistyakova_intellectual_2019} developed an intellectual system for the control of refractory lining, and \citet{dong_monitoring_2023} proposed a monitoring methodology for a converter process.
In terms of studies related to ladle metallurgy, \citet{wang_prediction_2018} developed an outlier detection scheme for the temperature of molten steel in the ladle furnace.
Similarly, \citet{emelianov_information_2022} also focused on monitoring the lining condition in the BOF, but their method relies on thermograms of steel ladles.
\citet{liu_anomaly_2021} proposed an anomaly detection method of the ladle driving force.
Finally, \citet{kim_development_2006} implemented a lifetime evaluation system for a ladle crane.
\citet{neto_deep_2021} proposed a method to optimize the maintenance policy of a shredder, a machine used to crush steel scrap before entering the steelmaking process.

\subsubsection*{Continuous Casting}
The studies devoted to the CCM often focused on monitoring the entire process~\citep{holloway_integration_1991,xu_multi-stage_2023,yang_process_2021,zhou_application_2022}.
Regarding the maintenance of individual equipment, the greatest attention was paid to mold monitoring~\citep{hutchison_recognition_2005,lukyanov_intelligent_2015,wu_anomaly_2023}.
Other equipment studied includes bearings~\citep{qing_liu_fault_2016,wu_multi-agent_2014}, casting roller~\citep{han_framework_2021}, crystallizer~\citep{david_j_and_svec_p_and_frischer_r_modelling_2012}, and exhaust fan~\citep{liu_toward_2023}.
Furthermore, \citet{zhou_online_2022} studied the CCM subsystem related to casting speed.

\subsubsection*{Hot Rolling Mill}
Hot rolling mill is the installation within the steel industry, which received the most attention from researchers.
For a detailed list of articles related to HRM with respect to the scope and type of asset considered, we refer the reader to Tab.~\ref{tab:hrm}.
The highest number of research papers was devoted to a finishing mill, which is a subsystem consisting of multiple rolling stands placed in tandem; its task is to reduce the hot strip to the desired thickness.
Within this subsystem, some studies were more focused and devoted to a single rolling stand~\citep{acernese_novel_2022,chen_fault_2023,lisounkin_advanced_2002,panagou_feature_2022,ruiz-sarmiento_predictive_2020}, work rolls~\citep{anagiannis_energy-based_2020,jakubowski_anomaly_2021,jiao_remaining_2021,kovacic_roll_2019,lebrun_mathematical_2013} or backup rolls~\citep{yuan_fatigue-damage_2023}.
Furthermore, the roughing mill was studied by \citet{ma_novel_2021-1}, which is a preliminary rolling stand that reshapes the steel slab before it enters the finishing mill.

Regarding work related to other hot rolling subsystems, we identified studies that address the problem of belt conveyor monitoring~\citep{junichi_t_and_yukinori_i_condition_2020}, laminar cooling~\citep{wang_integrated_2023}, and feedback control system~\citep{li_performance_2020}.
For the studies focused on individual devices, gearbox (see Tab.~\ref{tab:hrm})
, motor~\citep{fagarasan_signal-based_2004,ma_fault_2023,sarda_multi-step_2021}, looper\citep{garcia-beltran_causal-based_2004,jiang_optimized_2021,marcu_t_and_koppen-seliger_b_and_stiicher_r_hydraulic_2004,marcu_design_2008}, bearings~\citep{farina_fault_2015,pan_data-driven_2016,peng_multi-representation_2022} and pumps~\citep{goode_plant_2000} were also subjected to PdM tasks.
Finally, we identified many studies devoted to monitoring the entire installation or steel strip, which are listed in Tab.~\ref{tab:hrm}.

\subsubsection*{Pickling Line}
In the study of the pickling process, four articles exclusively devoted to this process were identified.
\citet{bouhouche_fault_2005,bouhouche_s_and_lahreche_m_and_ziani_s_and_bast_j_quality_2006} focused on fault detection in the loop control system of the production line.
\citet{fan_independent_2017} considered a waste liquor recovery plant, and
\citet{fleischanderl_cnnbased_2022} focused on defect detection on the strip surface.

\subsubsection*{Cold Rolling Mill}
Variety of different approaches were proposed for monitoring the cold rolling process.
The highest number of articles is related to the PdM of the entire production plant~\citep{beden_scro_2021,beden_towards_2023,chen_multi-source_2020,lepenioti_katerina_and_pertselakis_minas_and_bousdekis_alexandros_andlouca_andreas_and_lampathaki_fenareti_and_apostolou_dimitris_andmentzas_gregoris_and_anastasiou_stathis_machine_2020,wang_multi-step-ahead_2023}, which technologically resembles the HRM finishing mill.
Some effort of researchers is present in monitoring the strip~\citep{chen_multi-faceted_2021,yang_sw_and_widodo_a_and_caesarendra_w_and_oh_js_and_shim_mc_and_kim_sj_and_yang_bs_and_lee_wh_support_2009,yang_sliding_2021}, maintenance of work rolls~\citep{fouka_afroditi_and_bousdekis_alexandros_and_lepenioti_katerina_andmentzas_gregoris_real-time_2021,jakubowski_roll_2022,lakshmanan_data_2022,wang_sae-cca-based_2021}, rolling stand~\citep{lu_prediction_2020,shi_novel_2022,shin_k-y_and_kwon_w-k_development_2018} and tension system~\citep{arinton_neural_2012}.
In terms of individual equipment, we identified studies involving the detection or diagnosis of bearings~\citep{ji_fault_2021,shi_intelligent_2022,zhao_fault_2021} or motor~\citep{yanbin_sun_and_yi_an_research_2009} failures.
Finally, the study by \citet{lu_prediction_2020} focused on the problem of chattering, which is a common problem that occurs in CRM.

\subsubsection*{Annealing Line}
In the context of annealing, there exist two main technologies through which the process is conducted batch annealing and continuous annealing.
All of the surveyed papers are dedicated to the continuous annealing process.
\citet{yingwei_zhang_fault_2013}, \citet{liu_multiblock_2014,liu_quality-relevant_2014} and \citet{tiensuu_intelligent_2020} used measurements from the entire installation to monitor the process.
Additionally, Liu et al. studied tension subsystem~\citep{qiang_liu_data-based_2011} and strip thickness faults~\citep{liu_dynamic_2018}.
\citet{tan_shuai_and_wang_fuli_and_chang_yuqing_and_chen_weidong_and_xujiazhuo_fault_2010} focused their research on the annealing furnace, while \citet{wang_process_2012} and \citet{lu_pcasdg_2013} limited their methods only to the furnace temperature system.
Finally, \citet{zhang_fault_2011} addressed the problem of fault diagnosis in the looper.

\subsubsection*{Galvanizing Line}
There are two widely used techniques for galvanizing steel: \emph{Hot Dip Galvanizing} (HDG), which involves immersing the steel strip in a zinc bath and \emph{Electrogalvanizing}, which involves immersing it in electrolyte and applying electric current. 
In our review of the literature, we found only articles related to the HDG process. 
A commonly explored element in these studies is the welder, responsible for joining steel strips at the beginning of the process~\citep{bonikila_failure_2022,choi_modeling_2023,meyer_anomaly_2022}. 
\citet{debon_fault_2012} applied PdM methods to the zinc bath, \citet{simon_health_2021} specifically focused on the zinc bath roller, and \citet{chen_application_2021} studied bearing faults.
Finally, condition monitoring was applied to the entire galvanizing process by \citet{fan_application_2018}.

\subsubsection*{Miscellaneous}
Here we discuss all facilities and approaches that do not fit the installations discussed above or cover more than one production line.
The work of \citet{ray_artificial_1989}, which is the oldest identified study, is related to the PdM system for steel production facilities, in general.
An integrated approach was also presented in the work~\citet{cao_core_2022}, that proposed an ontology for the steelmaking process.
Other use cases, which incorporate multiple lines, were presented in works by
\citet{yanbin_sun_and_yi_an_research_2009-1} (cold rolling and pickling), 
\citet{shigemori_hiroyasu_quality_2013} (steel plant and hot rolling), 
\citet{slany_consistency_1996} (steel plant and CCM),
\citet{deepak_multivariate_2020} (EAF, steel plant, CCM and hot rolling),
\citet{kumar_nonlinear_2022} (sintering, blast furnace, steel plant and hot rolling) and
\citet{tiensuu_intelligent_2020} (hot rolling, cold rolling, pickling and annealing).
Moreover, the works of \citet{jin_varying-scale_2023,wu_anomaly_2021} were dedicated to the Energy Management System, which concerns the flows of energy gases between various installations.

Some of the researchers did not specify the steel facilities, which were examined in their work.
Three of these studies~\citep{hao_intelligent_2022,hutchison_approach_2014,straat_industry_2022} concern the monitoring of the steel strip.
Moreover, we identified studies that applied PdM to a common equipment present in steel factories but did not specify the installation, making it impossible to deduce it.
These studies covered maintenance of fan machinery~\citep{yi_jiangang_and_zeng_peng_analysis_2009}, air compressor~\citep{ganeha_b__k_n_umesh_condition_2019} and engine~\citep{ray_equipment_1991}.
    
\subsection{RQ2: What AI methods are used?}
We aimed at categorizing all the types of AI methods used in the articles. 
During the analysis of the methods employed, we identified certain studies that do not use strictly AI-based methods, for example, statistical or analytical models.
However, to make this survey more comprehensive, we decided to include them in the findings, as these methods can stand as an alternative to AI-based approaches.

\begin{figure}[ht]
    \centering
    \includegraphics[width=\columnwidth]{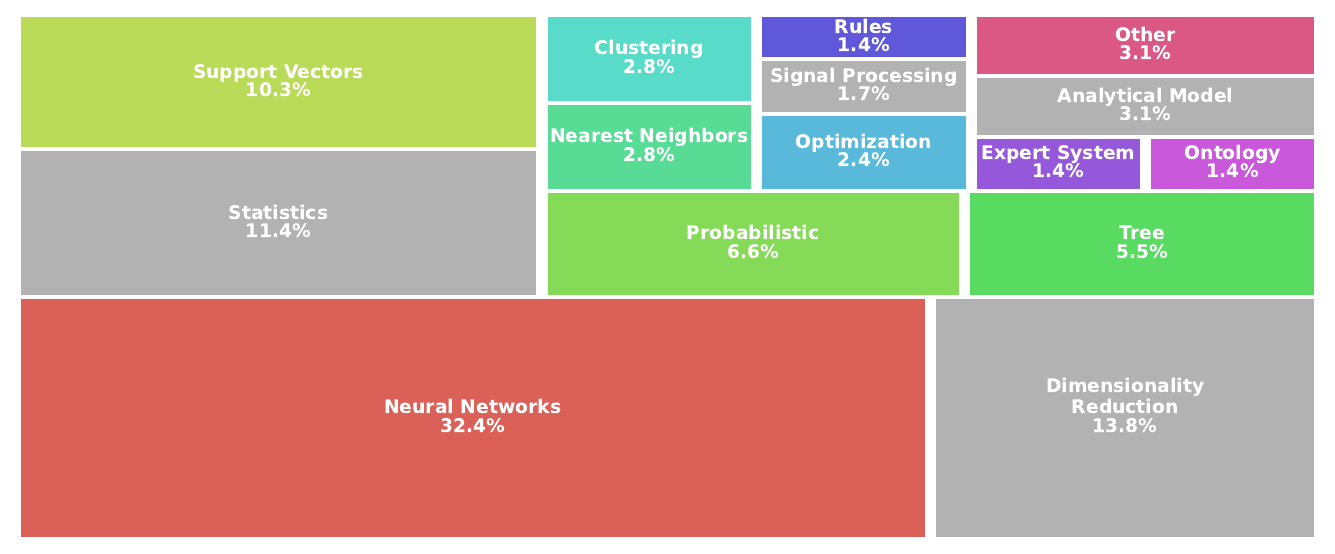}
    \caption{Occurrence frequency of methods used in the surveyed articles. Non-AI methods are colored in grey}
    \label{fig:treemap_ai}
\end{figure}

Fig.~\ref{fig:treemap_ai} presents the frequency of occurrence of each group within the analyzed articles.
We observe the predominance of ML methods, with NN being the most frequently one. 
Despite this, SVMs, probabilistic and tree-based methods are also widely used.
Non-AI approaches, such as statistics and dimensionality reduction, also had a significant share in the methods utilized.
We note that statistical approaches, which are discussed in more detail in Sect.~\ref{sec:statistics} were used mainly as primary prediction methods, but dimensionality reduction (discussed in Sect.~\ref{sec:dim_reduc}), was frequently used as preprocessing step for other ML methods.

\begin{figure}[ht]
    \centering
    \includegraphics[width=\columnwidth]{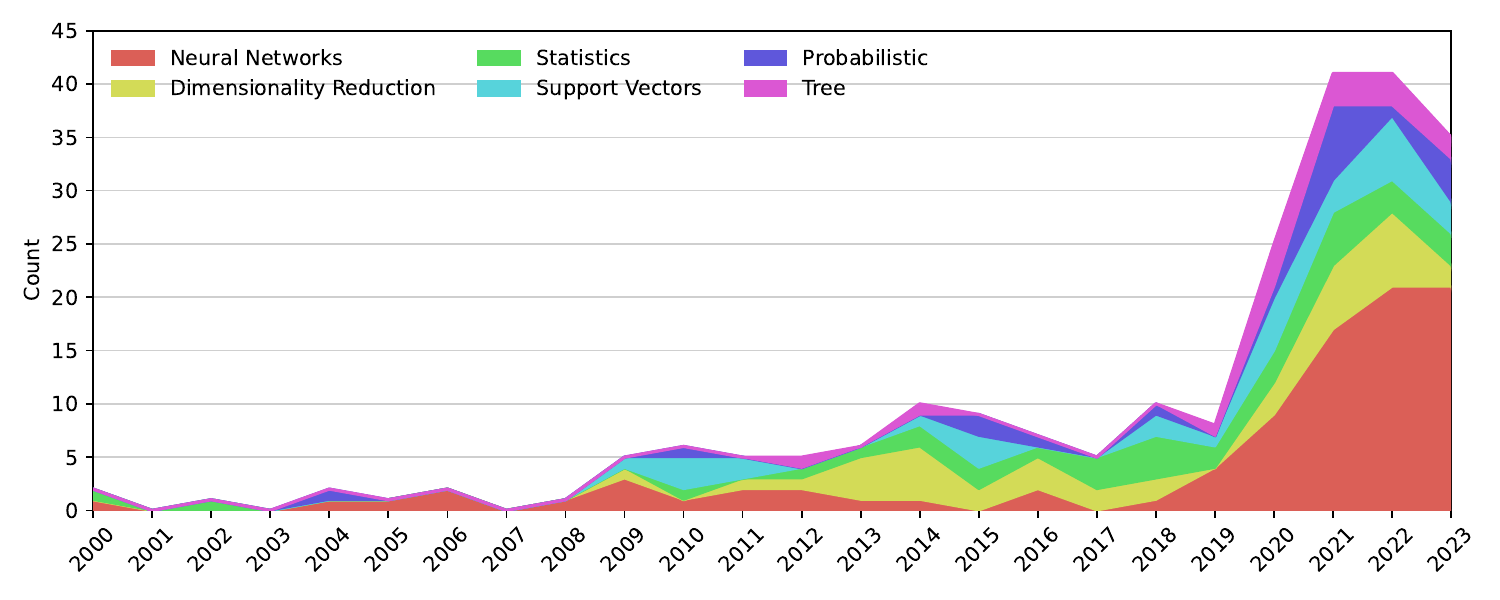}
    \caption{Temporal distribution of AI methods occurrence in surveyed articles. Articles using more than one method are counted multiple times, so the numbers are not consistent with total number of articles published each year.}
    \label{fig:ai_methods_timeline}
\end{figure}

Fig.~\ref{fig:ai_methods_timeline} presents historical trends with regard to the methods applied in the articles surveyed.
NN are one of the approaches that has appeared in the early stages of research in the surveyed domain and now are the predominant method employed.
The great increase in research in that area began around 2020, which we mainly associate with the development of deep learning methods, these are discussed in more detail in Sect.~\ref{sec:neural_networks}.

In the following subsections, we discuss each group of methods, highlighting the most common and original approaches.
While we briefly discuss the methods and algorithms utilized within each group, we focus on their application within the steel industry and do not delve into their mathematical foundations; for this, we refer the reader to specialized publications on this matter~\citep{Goodfellow_2016_deeplearning,Marsland2014_ML}.

\subsubsection*{Neural Networks}
\label{sec:neural_networks}
Within this group, we consider all types of neural network architecture, from multilayer perceptron (MLP) to deep learning methods.
Before we dive into the details of the findings, we want to clarify certain assumptions that were made to facilitate the analysis of methods based on neural networks.
We grouped similar NN approaches into larger groups; e.g. all autoencoder-based methods such as Variational AE, Sparse AE, Denoising AE etc. were categorized as "AE" and so on.
The only exception was in case of Recurrent Neural Networks -- due to the high number of articles related to this group of methods, we decided to treat Vanilla RNN, Long-Short-Term Memory (LSTM), and Gated-Recurrent Unit (GRU) networks separately.
If the paper did not specify exactly what type of architecture was used, we assumed that the regular MLP model was employed.
We observe that this is the case mainly in older publications and in ones that use relatively simple ML methods.

The results are summarized in Tab.~\ref{tab:neural}.
Although the highest number of studies use the MLP architecture, which is based on the perceptron model, many of these works date back to the 1990s and early 2000s~\citep{dinghui_zhang_researches_2000,marcu_t_and_koppen-seliger_b_and_stiicher_r_hydraulic_2004,meneganti_fuzzy_1998,ray_equipment_1991}.
The substantial growth of articles related to NN is observed from 2020 and is mainly related to research on more complex architectures such as RNNs, LSTMs, CNNs, or GRUs.
All these architectures are able to model complex temporal dependencies between the features, which make them particularly useful for time-series data, which are commonly used in the analyzed papers (for details, refer to Sect.~\ref{sec:rq4_datatypes}).
Additionally, CNNs are used in image recognition, so they can be applied to tasks that involve image processing~\citep{chernyi_application_2022,fleischanderl_cnnbased_2022,latham_tool_2023}.
Another wide group of methods are related to autoencoders, which can be used for unsupervised learning tasks such as anomaly detection~\citep{choi_modeling_2023,jakubowski_anomaly_2021,zhang_tongshuai_and_wang_wei_and_ye_hao_and_huang_dexian_and_zhanghaifeng_and_li_mingliang_fault_2016}.
In addition to the approaches discussed above, we observe some examples of less recognized strategies.
Several researchers applied the Generative Adversarial Network (GAN) together with other ML methods to predict faults in the production process\citep{peng_high-precision_2023,xie_fault_2021,zhang_robust_2023,zhao_fault_2021}.
Another example of a generative model, which is adapted to the PdM task, is the Deep Belief Network (DBN)~\citep{ji_fault_2021,jianliang_research_2020}.
We also identified some articles that used the attention mechanism in conjunction with CNN~\citep{hao_intelligent_2022,shi_intelligent_2022} and GRU~\citep{wang_research_2021}.
A notable example is the work of \citet{han_construction_2022}, who combined BERT~\citep{devlin2019bert} together with LSTM and the Graph Neural Network to construct a knowledge graph for HRM.
Another study, which uses the Transformer architecture, is the work of \citet{ma_fault_2023}.
Two authors employed the Double Deep Q Network~\citep{Hasselt2016ddqn} (DDQN), which is used mainly for reinforcement learning tasks; \citet{acernese_novel_2022} focused on fault detection in the rolling stand, while \citet{neto_deep_2021} used this approach to solve the maintenance scheduling problem.
Other examples of neural network architectures used in the surveyed papers include Restricted Boltzmann Machine~\citep{dong_monitoring_2023,liu_unsupervised_2023}, Radial Basis Function Network~\citep{perez_visual_2013} and Broad Learning System~\citep{zhang_extensible_2021}.

\subsubsection*{Support Vector Machines}
Algorithms in the SVM family can generally be grouped into three main approaches, depending on the type of ML task they solve.
Support Vector Classifier (SVC) is used for classification tasks, the Support Vector Regression (SVR) is applied for regression, and the One-Class SVM (OCSVM) is a commonly adopted method for outlier detection.
Except for OCSVM, some authors use the Support Vector Data Descriptor (SVDD) for one-class classification, but since these methods are closely connected and may be equivalent under specific conditions~\citep{xing_robust_2020}, we consider them as a single approach.
The synthesis of support vector-related papers is presented in Tab.~\ref{tab:suport_vectors}. 

In classification tasks SVM approaches are widely adopted, with some studies proposing extensions or modifications for improved performance~\citep{liu_multi-class_2011,tian_novel_2010,wang_anna_and_liu_zuoqian_and_tao_ran_fault_2010,wang_new_2015}. 
Most of these studies date back to the early 2010s. 
Another group of studies combines SVM with other methods~\citep{liu_optional_2011,lou_fault_2022,yang_sw_and_widodo_a_and_caesarendra_w_and_oh_js_and_shim_mc_and_kim_sj_and_yang_bs_and_lee_wh_support_2009}.
In regression approaches SVR was commonly used in comparative studies~\citep{lu_prediction_2020,song_application_2022,wang_multi-step-ahead_2023}.
\citet{zhang_fault_2021} used SVR in a semi-supervised manner to detect outliers, while \citet{liu_enhanced_2015} proposed a combination of SVM outlier detection with SVR to predict the quality of products in BF.

One-class classification methods based on SVM were often combined with other ML approaches~\citep{ouyang_h_and_zeng_j_and_li_y_and_luo_s_fault_2020,wang_prediction_2018,wang_detecting_2018,wang_blast_2022,zhai_multi-block_2020}. 
Furthermore, some articles utilized OCSVM in relatively simple scenarios, without any combinations with different methods or modifications of the original model~\citep{gerz_comparative_2022,russo_fault_2021,weber_learning_2022}.
In contrast, \citet{zhang_novel_2022} proposed probabilistic OCSVM for fault detection using SVM exclusively.
Finally, \citet{jiang_optimized_2021} used OCSVM as an outlier detector to find an optimal parity vector for a nonlinear system, which was later used as a fault detector. 

\subsubsection*{Probabilistic ML methods}
\label{sec:probabilistic}
In this group, Gaussian Mixture Models (GMM) were used in the highest number of studies, usually in conjunction with other algorithms~\citep{chen_hybrid_2023,han_novel_2022,peng_quality-related_2015,peng_dynamic_2021,xiao_novel_2023}, but also as a standalone algorithm~\citep{zhang_dynamic_2023}.
It should be noted that most of these studies have been published in the last 3 years, which implies a growing interest in GMMs.
Within the group of Gaussian-based approaches, a combination of the Gaussian process and SVDD was proposed for outlier detection~\citep{wang_prediction_2018}, as well as the Gaussian-based anomaly detection method for time-series~\citep{saci_autocorrelation_2021}.
Many studies were conducted that used Bayesian principles, combining them with other methods, i.e., dimensionality reduction~\citep{ma_novel_2021-1,peng_quality-based_2016} or neural networks~\citep{fouka_afroditi_and_bousdekis_alexandros_and_lepenioti_katerina_andmentzas_gregoris_real-time_2021,lian_fault_2010}.
Other approaches related to Bayesian inference include Bayesian Networks~\citep{lian_fault_2010}, Discrete Bayesian Filter~\citep{ruiz-sarmiento_predictive_2020} and Bayesian State Space Model~\citep{zhang_variational_2023}.
Bayesian inference was also used in~\cite{lu_pcasdg_2013} together with PCA for fault detection.
Finally, several studies showed the usability of Hidden Markov Models~\citep{sarda_multi-step_2021,simon_health_2021,wang_approach_2015}
and Causal Graph Models~\citep{garcia-beltran_causal-based_2004,yang_process_2021} for PdM tasks in the steel industry.

\subsubsection*{Tree-based}
\label{sec:tree_based}
The research related to tree-based algorithms was mainly focused on ensemble methods, that is, combining predictions of multiple weak learners to improve prediction accuracy and generalization.
A popular example of these methods is Random Forest, which was used in several studies for comparative evaluation with other algorithms~\citep{chen_multi-source_2020,chen_application_2021,hutchison_approach_2014,nkonyana_performance_2019,wang_multi-step-ahead_2023}.
Gradient Boosting Machines (GBM) offer a general improvement in the ensemble-based methods, compared to random forest.
Of these methods, XGBoost stands out as the most widely used algorithm. 
Analogically to Random Forest, it was used in comparative studies~\citep{lu_prediction_2020,wang_multi-step-ahead_2023}, but also as a standalone prediction algorithm~\citep{ding_high-precision_2023,wang_fault_2020}.
\citet{liu_online_2021} used LightGBM, which is another example of an algorithm for the GBM family.
\citet{tiensuu_intelligent_2020} also used a GBM-based model, but did not specify the details of the algorithm used.
More novel approaches, utilizing gradient boosting, were proposed by \citet{xie_fault_2021} and \citet{xu_novel_2024} -- both studies combined XGBoost with deep learning methods.
On the other hand, we also identified a few papers that utilized the single decision tree as a prediction algorithm~\citep{debon_fault_2012,hutchison_approach_2014,song_application_2022}.
The algorithms discussed above were used in classification and regression problems.
However, tree-based ensembles can also be used for outlier detection tasks, which was presented in the work by \citet{gerz_comparative_2022}, who used Isolation Forest along with other anomaly detection methods.

\subsubsection*{Clustering}
Clustering methods are generally used in tasks that require assembling similar observations into larger groups; however, within the surveyed papers only \citet{bo_qualitative_2015} and \citet{zhang_novel_2018} followed this approach.
We observe that the most popular direction of research was to apply clustering methods in
outlier detection tasks~\citep{gerz_comparative_2022,jin_varying-scale_2023,wang_detecting_2018,wu_anomaly_2021,zhao_adaptive_2014}.
Regarding the specific algorithms, the methods utilized in multiple studies inlucded k-Means~\citep{gerz_comparative_2022,shin_k-y_and_kwon_w-k_development_2018,wang_detecting_2018} and Adaptive Fuzzy C Means~\citep{wu_anomaly_2021,zhao_adaptive_2014}.
\citet{shin_k-y_and_kwon_w-k_development_2018} considered in their work also Mean-shifted clustering and density-based clustering.
These approaches used the corresponding algorithms in their standard form or combined them with other methods.
Some novelties were introduced in the work of \citet{zhang_novel_2018} who improved the Order-Preserving Submatrix algorithm, \citet{bo_qualitative_2015} who proposed the Qualitative Trend Clustering algorithm for fault diagnosis, and \citet{jin_varying-scale_2023} who proposed an algorithm based on DBSCAN for anomaly detection.

\subsection{Miscellaneous ML approaches}
Within this section, we include all ML methods that do not fit well into any of the groups discussed above.

Nearest Neighbors algorithm was widely used in conjunction with other methods.
Some authors combine this algorithm with other methods such as SVM~\citep{liu_optional_2011}, slow features analysis~\citep{dong_novel_2022}, PCA~\citep{zhang_pca-lmnn-based_2016}, clustering~\citep{zhao_adaptive_2014} and deep learning~\cite{ma_fault_2023}.
It is also used in simple comparative studies~\citep{fu_digital_2023,mentouri_steel_2020,wang_multi-step-ahead_2023}

From the group of linear models, we identified two studies using logistic regression~\citep{fu_digital_2023,shang_dominant_2017} and one dedicated to generalized linear models~\citep{debon_fault_2012}.
We note that methods such as PCA and PLS, which can also be categorized as linear models, were discussed in Sect.~\ref{sec:dim_reduc}.

Original approach is presented in works of \citet{ma_novel_2021,ma_novel_2022}, who proposed framework for classification of coupling faults, which they refer as Multi-task learning (MTL).
The method integrates multiple kernel learning and dimensionality reduction methods.

\subsection*{Knowledge-based methods}


In this section we briefly characterize minor, yet still important group of papers where knowledge-based techniques (in a broad sense) were used.
Rule-based expert systems were used in~\cite{miyabe_methodologies_1988} for supervision and control system,
then in~\cite{ray_artificial_1989} for maintenance management.
In~\cite{ray_equipment_1991} authors proposed a hybrid architecture where a high level rule-based component was used together with an MLP for fault diagnosis,
In~\cite{slany_consistency_1996} a consistency checking strategy for diagnostical rule-based system is also proposed,
and in~\cite{dinghui_zhang_researches_2000} a hybrid fault diagnostic system is described.
Most recently a hybrid fuzzy neural system was described in~\cite{ganeha_b__k_n_umesh_condition_2019} for condition monitoring.
In the early 21st century fuzzy rule-based systems where proposed for
anomaly detection in~\citet{lisounkin_advanced_2002},
and in~\cite{hutchison_recognition_2005} for failure prevention.
More recent papers using expert systems include~\cite{gao_intelligent_2010},
in~\cite{chistyakova_intellectual_2019} for maintenance management, and~\cite{spirin_expert_2020} for fault diagnosis.
In the last decade, ontologies where also used
with~\cite{zhang_knowledge_2012} for fault diagnosis,
\cite{beden_scro_2021} where a domain ontology for cold rolling was proposed, and then later used in~\cite{beden_towards_2023} where authors combine it with an ML classifier for maintenance scheduling.
In~\cite{cao_core_2022} a more general core reference ontology for steelmaking was described. 

\subsubsection*{Statistical methods}
\label{sec:statistics}
Statistics-based approaches are a distinct field of science, however they are the data-driven alternative to some of the ML approaches.
This fact, together with the high occurrence of statistics-based methods in the selected articles, encouraged us to include them in our analysis.
%
Independent Component Analysis, with certain modifications, was applied by \citet{tan_shuai_and_wang_fuli_and_chang_yuqing_and_chen_weidong_and_xujiazhuo_fault_2010} and \citet{fan_independent_2017,fan_application_2018}.
Other studies combined ICA with the PCA method~\citep{peng_new_2014,peng_new_kernel_2014,wang_process_2012,yingwei_zhang_fault_2013,zhou_data-driven_2021}.
Zhang and Xi et al. extensively applied Fractional Brownian Motion in their studies concerning the wall of BF~\citep{xi_remaining_2020,zhang_remaining_2017,zhang_fbm-based_2019,zhang_predicting_2019,zhang_remaining_2021}.
Slow Features Analysis was applied by \citet{zhai_multi-block_2020} and \citet{dong_novel_2022}.
Other statistical methods applied in the surveyed studies include
Canonical Correlation Analysis~\citep{liu_dynamic_2018,peng_new_2014,wang_sae-cca-based_2021}, Canonical Variates Analysis~\citep{lou_structured_2023}, 
Feature Selection~\citep{oh_jun-seok_and_kim_hack-eun_case_2015}, 
Least Squares~\citep{zhang_kpi-based_2017},
Statistical Trend Decomposistion~\citep{wu_anomaly_2021}, 
Principal Component Pursuit~\citep{pan_fault_2016,pan_robust_2018},
Mahalanobis Distance~\citep{an_graph-based_2020,shin_k-y_and_kwon_w-k_development_2018} and
Grey Model~\citep{zhao_anomaly_2022}.
Some non-standard statistical tests were also applied, including Scheffe test~\citep{lisounkin_advanced_2002} and Mann-Kandall test~\citep{weng_online_2023}.
We observed that statistical approaches were predominantly used for outlier detection tasks in an unsupervised approach.


\subsubsection*{Dimensionality reduction}
\label{sec:dim_reduc}
Similar to statistics, dimensionality reduction methods are not strictly an AI-based approach, but are widely used in data-driven studies to perform tasks such as preprocessing, visualization, or outlier detection.
The summary of studies that used the dimensionality reduction methods is presented in Tab.~\ref{tab:dim_reduc}.
The PCA is the dominant approach that was used in most studies concerning dimensionality reduction.
It was mainly applied together with other AI methods, generally as a preprocessing step (see Tab.~\ref{tab:dim_reduc}).
The researchers also combined PCA with other statistical methods, that is, ICA~\citep{peng_new_2014,peng_new_kernel_2014,wang_process_2012,yingwei_zhang_fault_2013,zhou_data-driven_2021}, CCA~\citep{peng_new_2014} and the Grey Model~\citep{zhao_anomaly_2022}.
Several studies used PCA, PLS, or their modifications alone (see Tab.~\ref{tab:dim_reduc}).
In general, in these studies, the authors reduce the dimension of the data and then compute the residual of each feature with respect to the projection, which is then used as a measure of the anomaly; based on the value of the squared prediction error or the test $T^2$, outlier detection is performed.
Furthermore, \citet{peng_quality-related_2015,peng_quality-based_2016,zhang_extensible_2021} and \citet{chen_application_2021} combined PLS with other AI methods, e.g. probabilistic models, trees, or statistical methods.
Except for PCA-based and PLS-based studies, we identified only two studies that used other dimensionality reduction methods.
\citet{perez_visual_2013} used t-distributed Stochastic Neighbor Embedding (t-SNE) with RBF Neural Network, and \citet{sha_robust_2022} utilized Neighborhood Preserving Embedding (NPE).

\subsubsection*{Other techniques}
Within this section we describe miscellaneous methods, which do not fit into subfields described in the above sections.

The first group of methods is related to optimization techniques.
Most of the research on these methods was related to hyperparameter tuning or improving the learning process of the ML model.
Several studies~\citep{ji_fault_2021,jing_intelligent_2023,qing_liu_fault_2016} used Genetic Algorithms (GA) to find the optimal parameters of ML models, \citet{zhang_fault_2021} used the Chaotic Sparrow Search Algorithm to optimize the parameters of the SVM, Jakubowski et al. used Bayesian optimization to tune the hyperparameters of the autoencoder in two of their studies~\citep{jakubowski_anomaly_2021,jakubowski_roll_2022}, and \citet{yang_comparison_2015} compared GA with Particle Swarm Optimization (PSO) and grid search to tune the SVM model.
Regarding other applications of optimization methods, \citet{wang_shuqiao_and_yuan_yan_and_liang_jingjie_and_zhang_zhuofu_design_2022} utilized PSO to improve the efficiency of the analytical model, and \citep{kovacic_roll_2019} used Genetic Programming to find the optimal mathematical model of roll wear.

Analytical models are the next category, which does not fall strictly into the group of AI methods. 
The objective of these models is to simulate the behavior of the subject of study with the means of physics-based equations and knowledge of the underlying processes.
We find that many of the identified studies, utilizing analytical models, were related to prediction of roll wear in rolling mills~\citep{anagiannis_energy-based_2020,jakubowski_roll_2022,lebrun_mathematical_2013,yuan_fatigue-damage_2023}.
Moreover, a wider scope of the hot rolling process was studied by \citet{lisounkin_advanced_2002} and \citet{ball_model_2020}.
Another HRM use case is a study by \citet{wang_integrated_2023}, who used the finite element method to monitor the laminar cooling process.
Studies in other steel production facilities included monitoring of hearth erosion of BF~\citep{wang_shuqiao_and_yuan_yan_and_liang_jingjie_and_zhang_zhuofu_design_2022} and metal breakages in CCM~\citep{lukyanov_intelligent_2015}.
Some of the discussed approaches, in addition to using an analytic model, also utilized AI methods.
Lisounkin~\citep{lisounkin_advanced_2002} used a fuzzy expert system to detect faults in the HRM,
\citet{wang_shuqiao_and_yuan_yan_and_liang_jingjie_and_zhang_zhuofu_design_2022} used PSO for optimizing their algorithm,
\citet{kim_development_2006} created finite-element simulation of ladle crane fatigue, which was later used to train neural network,
and \citet{jakubowski_roll_2022} injected an analytical model of roll wear into the autoencoder.

In addition, we identified some studies, which used strictly signal processing methods such as Fast Fourier Transform~\cite{dehghan_marvasti_fault_2014,fagarasan_signal-based_2004,shiyan_vibration-based_2019} or Wavelet Transform~\cite{pan_data-driven_2016,sun_research_2012}.
Although these are not AI-based methods, they can serve as data preprocessing step and are valuable tools, especially in case of high-frequency signals.


\subsubsection*{The use of Explainable AI methods}
The analysis of the papers with regard to the use of interpretable or explainable models showed that only a small number of studies addressed this issue.
We observe that there are two leading approaches to this matter.
The first group used contribution plots, which illustrate the impact of individual features on the prediction for a specific instance or observation in the dataset.
These plots were adopted in several studies that used statistical methods~\citep{guo_quality-related_2023,kumar_nonlinear_2022,liu_structured_2018,zhang_fault_2011,zhao_anomaly_2022} and ML-based approaches~\citep{han_novel_2022}.
Second group utilized the SHAP~\citep{lundberg2017} method, which determines the contribution of each feature to the model output based on Shapley values from cooperative game theory.
This method was applied to explain the prediction of the deep learning~\citep{jakubowski_anomaly_2021,jakubowski_explainable_2021} and Gradient Boosting~\citep{ding_high-precision_2023,tiensuu_intelligent_2020} methods.
In addition, counterfactual explanations were utilized in one study~\citep{jakubowski_roll_2022}, in which an autoencoder model was developed.

\subsection{RQ3:  What types of PdM approaches are used?}

The analysis of papers with respect to the type of PdM approach followed is difficult due to inconsistencies in the terminology used by the authors.
Some authors do not clearly use the common terminology, which we define in Sect.~\ref{sec:pdm}, but rather define the PdM task using their own or less recognized terms.
Other common issues 
include the usage of some terms interchangeably or an imprecise definition of the PdM task employed.
To maintain objectivity, we have strictly adhered to the terminology used by the authors, but in some cases (in which terminology was not explicitly used) the type of task was inferred based on the provided information.
However, it is important to note that inconsistencies persist. 

\begin{figure}[ht]
    \centering
    \includegraphics[width=\columnwidth]{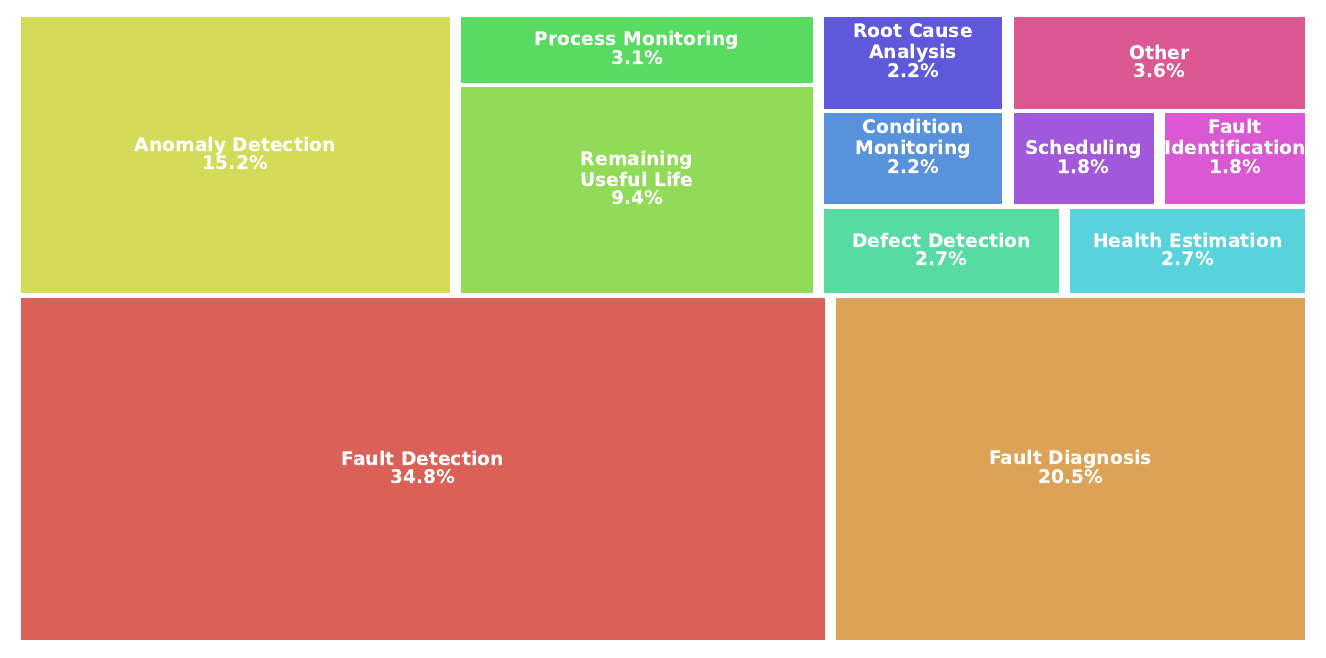}
    \caption{Frequency of occurrence of different PdM tasks within the surveyed papers}
    \label{fig:pdm_tasks}
\end{figure}

Fig.~\ref{fig:pdm_tasks} provides an overview of the frequency of different PdM tasks. 
Most research articles focus on diagnostics, such as fault detection, fault diagnosis, and anomaly detection, comprising approximately 70\% of the research output. 
These tasks represent fundamental concepts within the diagnostics domain.
For a complete list of papers using these approaches, we refer the reader to Tab.~\ref{tab:pdm_tasks}.
We observe certain ambiguities in the terminology used by various authors in describing these tasks. 
In our opinion, this ambiguity arises from the fact that these methods share overlapping characteristics, leading to subjective definitions of tasks. 
For example, some studies employ an autoencoder architecture with reconstruction error, which is described as anomaly detection~\citep{jakubowski_explainable_2021,meyer_anomaly_2022,zhou_application_2022} or fault detection~\citep{lu_prediction_2020,zhang_tongshuai_and_wang_wei_and_ye_hao_and_huang_dexian_and_zhanghaifeng_and_li_mingliang_fault_2016,zhang_robust_2023} in different works. 
Similarly, some authors use the terms \emph{fault detection} and \emph{fault diagnosis} together to describe their methods~\citep{deepak_multivariate_2020,lisounkin_advanced_2002,russo_fault_2021}.

Several authors claimed that their method applies more in-depth diagnostic tasks than fault diagnosis.
\citet{liu_toward_2023} proposed a fault identification scheme based on autoencoders, which was applied for CCM.
\citet{zhou_fault_2020,zhou_data-driven_2021} used PCA-based and PLS-based algorithms to calculate the deviation of the BF process from normal working conditions, then fault identification was performed by estimating the contribution rate of each feature to this deviation.
\citet{zhang_comprehensive_2023} proposed a framework to assess the operating level and identify HRM faults based on the GRU network.
\citet{zhang_extensible_2021} also proposed a fault isolation method using PLS and the Broad Learning System to differentiate between different types of fault in HRM.
\citet{fagarasan_signal-based_2004} developed a diagnostic agent that performs fault detection and isolation based on the torsional vibration of a main drive in HRM.

Root Cause Analysis, being the most comprehensive diagnostic task, has also gained some attention from researchers.
\citet{tiensuu_intelligent_2020} proposed a system to discover the root cause of center line deviation during steel strip processing, which is based on the GBM and SHAP methods.
\citet{han_novel_2022} used the graph neural network to diagnose faults in the BF process.
The inference mechanism was then built using contribution plots.
\citet{kumar_nonlinear_2022} studied fault diagnosis in the integrated steel factory, which was based on the neural network and PCA, the root cause analysis of the faults was also performed using contribution plots.
\citet{ma_practical_2022} proposed a method for the analysis of the root cause in the HRM finishing mill, which uses the GRU network and the Granger causality.
Lastly, \citet{latham_tool_2023} developed a tool, which combines expert knowledge and ML methods to determine the root causes of width-related defects in the hot rolling process.

Process monitoring was explored in many studies, the methods within this group focused mainly on the outlier detection, making them similar to anomaly detection tasks.
These studies were devoted to monitoring the entire BF process~\citep{chen_hybrid_2023,liu_structured_2018,zhang_novel_2018}, and the finishing mill in the HRM~\citep{peng_new_2014,peng_new_kernel_2014,zhang_comprehensive_2023}.
In some of them, the monitoring of the process was also performed in conjunction with fault detection~\citep{liu_structured_2018,peng_new_2014} or fault identification~\citep{zhang_comprehensive_2023}.
Condition monitoring was used in some articles and referred to tasks such as rolling stand monitoring in CRM~\citep{oh_jun-seok_and_kim_hack-eun_case_2015}, torpedo car in BF~\citep{yemelyanov_application_2021}, hearth erosion in BF~\citep{wang_shuqiao_and_yuan_yan_and_liang_jingjie_and_zhang_zhuofu_design_2022} and transformer in EAF~\citep{shiyan_vibration-based_2019}.

Defect detection, which monitors the degradation of the product, rather than the equipment, was devoted only to the steel strip.
\citet{yang_novel_2023} use visual data from the HRM monitoring system to detect defects, while \citet{fleischanderl_cnnbased_2022} work with data from the pickling line.
Two of the works on the recognition of defects in steel strip did not specify the manufacturing site~\citep{hao_intelligent_2022,hutchison_approach_2014}.
\citet{song_application_2022} and \citet{liu_enhanced_2015} focused on the quality prediction of the product in HRM and BF, respectively, which is closely related to defect detection.

The number of articles considering prognostics approach is much smaller than in the case of diagnostics.
Within this group, most of the studies were related to the estimation of RUL, which represents approximately 10\% of the articles.
\citet{zhang_remaining_2017,zhang_predicting_2019,zhang_fbm-based_2019,zhang_remaining_2021} and \citet{xi_remaining_2020} extensively explored this topic, focusing on predicting RUL of blast furnace walls using fractional Brownian Motions.
In one of the works, \citet{zhang_intelligent_2022} also used RNN for a similar task.
Within the BF area, \citet{raducan_prediction_2020} developed a method for RUL estimation of blowers; however, no details on the method were provided.
Several studies were also dedicated to CCM, which considered bearings~\citep{qing_liu_fault_2016,wu_multi-agent_2014}, casting roller~\citep{han_framework_2021}.
Moving on to HRM, the RUL of the work rolls was a significant research area~\citep{anagiannis_energy-based_2020,jiao_remaining_2021,kovacic_roll_2019}, but also a complete finishing mill~\citep{ball_model_2020}.
In addition, \citet{goode_plant_2000} aimed to predict the time for the next pump maintenance.
Other studies considering the prediction of RUL were dedicated to CRM~\citep{lakshmanan_data_2022,lepenioti_katerina_and_pertselakis_minas_and_bousdekis_alexandros_andlouca_andreas_and_lampathaki_fenareti_and_apostolou_dimitris_andmentzas_gregoris_and_anastasiou_stathis_machine_2020}, crane in steel plant~\citep{kim_development_2006}, HDG line~\citep{simon_health_2021}, and EAF~\citep{choi_method_2020}.
Finally, \citet{karagiorgou_making_2020} developed the LSTM-based approach for the prediction of RUL; however, they did not specify the exact steel facility under study.

The studies, which were aimed at estimating the HI, were primarily related to the monitoring of individual devices.
Many of these studies considered equipment in CCM, that is, bearings~\citep{qing_liu_fault_2016,wu_multi-agent_2014} and crystallizer~\citep{david_j_and_svec_p_and_frischer_r_modelling_2012}.
Other works were dedicated to rolls in rolling mills; more precisely, to backup rolls in HRM~\citep{yuan_fatigue-damage_2023}, work rolls in HRM~\citep{lebrun_mathematical_2013} and the work rolls in CRM~\citep{fouka_afroditi_and_bousdekis_alexandros_and_lepenioti_katerina_andmentzas_gregoris_real-time_2021}.
\citet{ruiz-sarmiento_predictive_2020} also tried to predict the health condition of the asset, in this case the equipment in Steckel HRM, but referred to it as a prediction of \emph{degradation state}.
Similarly, \citet{chen_application_2021} do not explicitly use the term HI, but their method, which predicts bearing wear, is equivalent to calculating HI.

The research on scheduling tasks, which seems to be an important aspect of PdM from a business perspective, was very limited.
\citet{ray_artificial_1989} developed an expert system for a steel industry, which generates maintenance rules based on the knowledge base.
It does not provide automatic scheduling; however, it supports engineers in scheduling proper maintenance tasks.
\citet{holloway_integration_1991} proposed an integrated approach for fault detection and reactive production scheduling in CCM.
We note that these papers date back to around 1990. 
From more recent work, \citet{neto_deep_2021} developed a reinforcement learning method to optimize the maintenance schedule of the shredder in the steel plant.

\subsection{RQ4: What are the characteristics of the data used? }
\label{sec:rq4}

In AI research, the characteristics of the data are one of the most important factors that determine the feasibility of the proposed approach.
Within this research question, our aim is to determine these characteristics with respect to three factors.
First, we analyzed the data types used for PdM tasks in the steel industry.
Then, we assessed the source of the data, with the focus on its authenticity, that is, whether it is an artificial or real-world data set.
Lastly, we inspect the availability of the data to other researchers, which is one of the crucial aspects supporting the reproducibility and comparison of the AI methods.

\subsubsection*{Data types}
\label{sec:rq4_datatypes}
\begin{figure}[ht]
    \centering
    \includegraphics[width=\columnwidth]{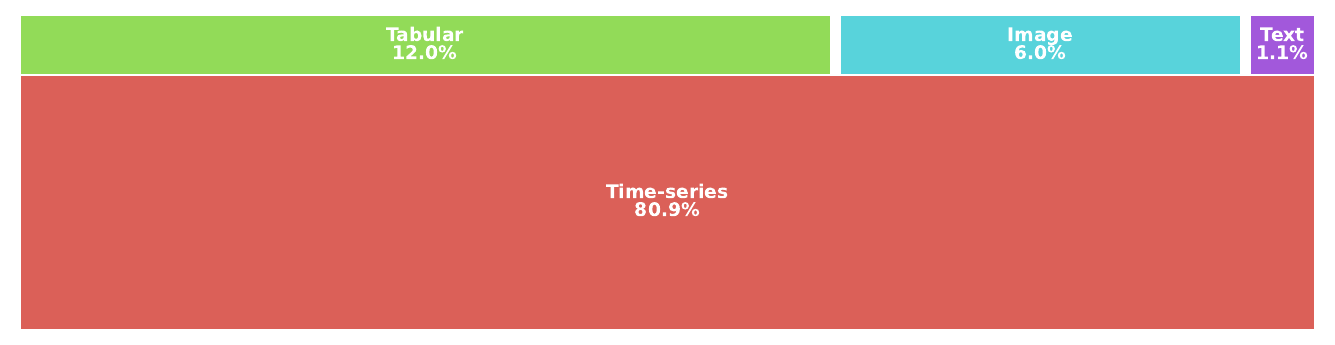}
    \caption{Frequency of using different data formats by researchers}
    \label{fig:data_formats}
\end{figure}

The type of data used plays an important role in determining the optimal AI method and the PdM approach.
The volume and quality of the data have a high impact on the performance of the model and can restrict the selection of potential methods.
Fig.~\ref{fig:data_formats} presents the frequency of using different types of data in the articles surveyed.
We identified four types of data, which were used: time series, tabular, image, and text.

The primary type of data used is time series, where the information is generated sequentially over time.
This mainly involves data from sensor readings that are installed in the facilities. 
Such data can be easily processed and stored in the form of files or databases.
We observe that algorithms that are able to learn temporal relationships are often used for this task.
This includes mainly deep learning approaches, e.g., LSTM Networks~\citep{jiao_remaining_2021,karagiorgou_making_2020,lakshmanan_data_2022,lepenioti_katerina_and_pertselakis_minas_and_bousdekis_alexandros_andlouca_andreas_and_lampathaki_fenareti_and_apostolou_dimitris_andmentzas_gregoris_and_anastasiou_stathis_machine_2020}, GRUs~\citep{li_abnormality_2023,wang_research_2021,wu_anomaly_2023}, CNNs~\citep{gao_minimax_2023,liu_anomaly_2021,peng_multi-representation_2022}, but also statistical~\citep{zhang_remaining_2017,zhang_fbm-based_2019,zhang_predicting_2019} and probabilistic~\citep{saci_autocorrelation_2021} approaches.
Alternatively, time-series data can be treated similarly to tabular data, i.e., assuming no temporal dependency between the observations.
In such scenarios, popular ML approaches such as Gradient Boosting~\citep{liu_online_2021,wang_multi-step-ahead_2023}, SVM~\citep{konishi_masami_and_nakano_koichi_application_2010,liu_enhanced_2015,russo_fault_2021} can be utilized, but we lose the temporal information, which might be relevant for certain use cases.

Tabular data was also adopted, as similarly to time-series, it comes primarily from sensor readings.
Usually, the difference is that it contains aggregated measurements for a certain product~\citep{debon_fault_2012,ding_high-precision_2023,jakubowski_roll_2022,nkonyana_performance_2019,tiensuu_intelligent_2020} instead of raw sensor readings.
Except for product-based measurements, approaches that use tabular data include sampling of measurements to create a balanced data set~\citep{liu_optional_2011,wang_anna_and_liu_zuoqian_and_tao_ran_fault_2010}, using recorded maintenance data~\citep{kovacic_roll_2019} or fusing information in tabular form with other data structures~\citep{jakubowski_explainable_2021,meyer_anomaly_2022,wang_fault_2020}.

Several studies used visual data in the form of images to develop PdM solutions.
These studies are heavily based on the use of CNN architecture to process the data.
Furthermore, they are predominantly related to inspection of the steel strip surface~\citep{fleischanderl_cnnbased_2022,hao_intelligent_2022,jing_intelligent_2023,mentouri_steel_2020,peng_high-precision_2023,yang_novel_2023}.
Another source of image data is recordings from cameras, which are installed in critical areas of the facilities.
This approach was applied to torpedo cars~\citep{chernyi_application_2022,li_research_2020}, BF raceway~\citep{puttinger_improving_2019} and HRM ~\citep{junichi_t_and_yukinori_i_condition_2020}.
Despite the images from regular cameras, which were described above, an interesting possibility is to use images containing other than strictly visual information.
Some examples include an infrared camera, which produces a thermogram of the observed object~\citep{emelianov_information_2022} and measurements from a 3D scanner equipped with an ultrasonic probe~\citep{bonikila_failure_2022}.
A completely different approach was adopted by \citet{latham_tool_2023}, who converted time-series data into images.

Lastly, data in the form of text can also be used, although the research within this area is very limited, as we identified only two studies that addressed it.
The work of \citet{han_construction_2022} uses various deep learning architectures to build a knowledge graph for fault diagnosis based on the reports of workshop maintenance personnel.
\citet{wang_fault_2020} combined text and tabular data to detect faults related to steel surface quality; the textual description of the defect was used to determine the labels for a classification model.

\subsubsection*{Data sources}
With respect to the authenticity of the data, we observe that most of the studies were conducted on real measurements from a steel manufacturing facility.
Several authors also used synthetic data (not necessarily related to the steel industry) for the sake of performance assessment~\citep{garcia-beltran_causal-based_2004,han_novel_2022,jakubowski_anomaly_2021,lou_data-driven_2023}.
We also identified studies in which the actual data from the installations was used to train the model, but the validation was performed on synthetically manipulated data~\citep{pan_fault_2016,xie_fault_2021,zhang_fault_2021}.

The use of synthetic data was also adopted in many studies.
This involves, for example, simulated data coming from mathematical model~\citep{chen_fault_2023,konishi_masami_and_nakano_koichi_application_2010,miyabe_methodologies_1988}, digital twin~\citep{han_framework_2021} or image data which do not come from an actual installation~\citep{fleischanderl_cnnbased_2022,peng_high-precision_2023,yang_novel_2023}.
The alternative approach to utilization of data from real production line is to conduct a laboratory experiment and use these data for research.
This approach was adopted mainly for a cold rolling process~\citep{ji_fault_2021,shi_intelligent_2022,shi_novel_2022,wang_multi-step-ahead_2023}.

\subsubsection*{Availability of the data used}
To complete the analysis on the characteristics of the data used, we determined its availability to other researchers.
We did not find any work making the utilized data publicly available as an appendix to the published paper.
\emph{We believe that main reason for that may be the confidentiality of the data coming from the private steel companies.}
In total, we found three publicly available datasets, which are dedicated to steel production and were used by the authors of the articles surveyed.
UCI Steel Plates Defects~\citep{misc_steel_plates_faults_198} contains tabular data with measurements for a particular product together with the corresponding defect class and was applied to validate rather simple ML approaches~\citep{hutchison_approach_2014,nkonyana_performance_2019}.
NEU Surface Defect Database~\citep{Song2013_neu,neu_dataset} and the Severstal Steel Defect Detection Dataset~\citep{severstal_dataset} both contain images with steel surface defects.
They were used to show the effectiveness of deep learning methods for defect detection in steel plates~\citep{hao_intelligent_2022,mentouri_steel_2020,peng_high-precision_2023,yang_novel_2023}.
Finally, \citet{wang_multi-step-ahead_2023} used data from a laboratory experiment, which can be shared on reasonable request.

\subsection{RQ5: What is the business impact of the proposed methods?}
The business perspective of research in the steel industry plays a crucial role in the development of new methodologies and the improvement of existing ones.
The verification of the proposed methods in industrial scenarios, where a significant number of difficulties might be observed, is invaluable.
First, it ensures that the proposed methods are operational in the real installation and can tackle difficulties such as real-time prediction capability, concept drifts, or low data quality.
Second, the final objective of PdM methods is to reduce downtime, costs, and increase asset safety and reliability.
From such a perspective, the evaluation of the business gains for the proposed methods seems like an imperative step.

To measure the business impact of the articles surveyed, we defined three distinct criteria.
First, we checked if there was any sign of collaboration with the industry.
Some examples of collaboration include
1) the authors stated that they worked together with plant engineers or used their expert knowledge, 
    2) at least one author was affiliated to a steel-related company, and
3) the name of company was used within the text of the paper.
This means that using the data from a real production facility does not imply that there was any collaboration between the authors and the company, which shared the production data.
Second, we verified that the proposed solution was actually implemented in the real production environment.
This criterion is more strict than the first one and requires the authors to clearly indicate that the method they proposed is actually operating in the facility.
Finally, we checked whether the authors gave a quantitative estimate of the business benefits of the implementation of their method.
This criterion does not oblige the method to be implemented in the production facility, so if rough estimates based on reasonable assumptions are given, the criterion is still fulfilled.
The summary of the articles that met at least one of the criteria is given in Tab.~\ref{tab:rq5}.

Despite the decent number of papers that showed evidence of collaboration, only a limited number of papers highlight the successful implementation of the method.
\citet{holloway_integration_1991} are the first researchers identified who implemented an AI-based PdM solution for a steel factory.
Their work was dedicated to CCM, and a prototype of the behavioral model monitoring system was implemented on a MicroVax II computer.
\citet{lisounkin_advanced_2002} developed an analytical model combined with the fuzzy expert system to detect anomalies in the selected elements of HRM rolling stands.
\citet{garcia-beltran_causal-based_2004} developed a causal graph model for the hydraulic looper, the model runs in a dedicated application.
\citet{bouhouche_fault_2005} implemented a fault detection system on a process computer for a loop control system on the pickling line.
\citet{hutchison_recognition_2005} used fuzzy logic to prevent breakdowns in CCM.
The system is operating in a production environment and replaced an older solution.
\citet{kim_development_2006} developed a Web-based Enterprise Resource Planning system to monitor the condition of the ladle crane. 
\citet{shigemori_hiroyasu_quality_2013} proposed a quality monitoring system for various steel production processes, which was operated for c.a. 3 years before the results were published.
\citet{lukyanov_intelligent_2015} implemented an analytical model for the prediction of liquid metal breakouts in CCM, which operated for more than a year.
\citet{shin_k-y_and_kwon_w-k_development_2018} developed a smart monitoring system to diagnose the CRM condition and implemented it in the form of a Human-Machine Interface, allowing real-time display based on sensor data.
\citet{tiensuu_intelligent_2020} implemented a Web-based tool for online monitoring and root cause analysis for strip position deviation in HRM.
\citet{raducan_prediction_2020} applied ML methods to predict the RUL of turbo blowers in BF, the solution was implemented as a PowerBI application.
\citet{chen_application_2021} applied PLS for bearing wear prediction and implemented it as a web application.
\citet{wang_shuqiao_and_yuan_yan_and_liang_jingjie_and_zhang_zhuofu_design_2022} designed an analytical model that monitors the erosion process in the hearth of BF.
The results of the model are displayed to the plant staff to optimize the production process and maintenance strategy.
\citet{peng_dynamic_2021} implemented a graphical user interface to monitor strip thickness faults in HRM.
\citet{panagou_explorative_2022} developed a digital twin-based PdM solution, which was implemented as a web platform.
\citet{jin_varying-scale_2023} proposed an ML method for anomaly detection and applied it at the energy management center of the steel company.
Furthermore, \citet{han_framework_2021} declare that they have implemented a prototype based on digital twin for CCM; however, there are no details on collaboration with any specific steel enterprise.

In the context of business gains from the proposed methods, only a few studies were identified, which showed evidence of a quantitative evaluation.
\citet{lukyanov_intelligent_2015} reported the performance of their method after implementation: within 1.5 years of operation, the system detected 44 probable cases of failure while significantly reducing the number of false alarms, compared with the regular diagnostic system.
\citet{han_framework_2021} claim that their solution, based on digital twin, which predicts RUL for a casting roller in CCM, improves service time by 10-15\% and increases steel production by 23\%.
\citet{choi_modeling_2023} evaluated the financial costs of laser welder failure in HDG.
Based on the duration of stoppage, the productivity of line, and assumed hourly profit, they estimated that the model can help avoid a loss of about 23,000 USD per year.
This value can increase to 400,000 USD / year if the method is applied to other installations of the collaborated enterprise.

\section{Discussion}
\label{sec:discussion}

The analysis of the articles included in the survey can help us determine the current stage, gaps and perspectives in the research area of AI-driven PdM in the steel industry.

\subsection{Summary of the findings}
Below, we present the list of the main findings based on the results of our survey.
We believe that they give a global picture of current state of knowledge in the field.

First of all, the number of papers published in recent years has been growing constantly, which implies increased interest of the scientific community in the field.
%
Research on steel facilities spans the whole production pipeline, from the sinter plant to the galvanizing lines.
    However, a major research activity is observed in the hot-rolling and BF processes.
%
 The scope of research ranges from the monitoring of single devices to the whole production process.
    The equipment, which is subject of study varies from general-purpose devices, e.g. bearings, gearboxes, to industry-specific equipment like loopers and work rolls.

%
The terminology used to describe PdM tasks is often vague and certain types of tasks have overlying characteristics.
    This impedes the analysis of the papers with respect to the type of task addressed.
%
The highest number of papers are related to the prediction of faults with varying degree of profoundness, ranging from simple detection to more in-depth root cause analysis.
%
Examples of more unconventional PdM approaches include the estimation of the health state or remaining useful life, and the scheduling of maintenance actions.

As of analysis methods used,
most of the work is devoted to the development of ML methods to perform various PdM tasks, the presence of other AI methods is very limited.
%
A relatively high number of papers use unsupervised techniques, which do not require labeled data during training.
    We believe that the main causes of such situation are the high imbalance of data (between faulty and normal observations) and the lack of high-quality labels.
%
The increased amount of research in recent years is mainly dedicated to the development of deep learning methods.
%
There has been also a small number of papers where knowledge based technologies, mainly expert systems, were used.
However, besides recent work on the use of formal ontologies, this trend is of a marginal importance.
%
Finally, even though XAI has been a hot topic in AI for almost e decade, so far little effort has been made to provide explanations for model decisions using XAI techniques.
    
Data treatment is another important aspect to observe.
Most of the papers are based on data collected from sensors installed in the facilities.
    However, often a temporal dependency between the observations is not taken into account (data treatment is similar to tabular data).
%
We noted that several methods were proposed to deal with image and text data, which might be a promising research direction given the development of deep learning methods for computer vision and natural language processing.
%
%
%
 Most of the researchers use private data, which comes from a real steel production facility.
    The use of such data enhances the usefulness of the proposed methods in practice but also hinders the possibility of reproducing the results and comparing them with alternative approaches.
%
 There is very little public data available for researchers to benchmark their solutions in steel industry use cases.
    We identified only three such datasets, two of which are related to image processing and one being relatively simple tabular data.
%
Moreover,
the authors, with only one exception, do not provide the source code produced to conduct research.
    The proposed methods are therefore difficult to reproduce and validate, which is even more difficult taking into account the lack of research data.

%
The research is mainly devoted to existing installations, whose location is usually not revealed in the paper. 
    However, based on the available data, we observe that the location is generally aligned with the affiliation of authors.
%
Chinese universities occupy a dominant position in terms of the number of articles published.
    Significant research activity is also present in developed countries, especially in western Europe.
    This generally aligns with global steel production data~\citep{WorldSteel2023}.
%
Two larger clusters of researchers from Chinese universities conducted exceptionally high research activity.
    Except for these two cases, all other research teams worked independently.

Finally,
while the collaboration of researchers with companies is clearly visible, the observable results of these works are very limited.
    Only a limited number of studies declare that their method was implemented in the production environment.
    The quantitative benefits were presented in very few papers.


\subsection{Gaps and future directions}
While research on the application of AI methods for PdM in the steel industry has been extensive and has explored various methods and facilities, there remains ample room for improvement. 
Through our analysis, we have identified current trends and gaps that we believe should guide future research. 
In the following, we highlight the most important directions that should be considered by both academia and businesses.

\begin{itemize}
    \item At present, substantial research efforts are directed towards BF technology. 
    However, according to existing regulations and the environmental objectives outlined by the major steel producers~\citep{EuropeanUnion2021_zeroemission,Yang2020_EAF_LifeCycle,Yu2022_China_Zero}, it is expected that BF undergoes gradual decommissioning in favor of less emission-intensive processes such as EAF. 
    Consequently, we project an increase in interest towards PdM within the EAF process.

    \item The major growth of research in deep learning will probably continue, as the new architectures and methods offer high versatility and efficiency, which can encourage researchers to continue studies in this direction.
    Additionally, significant advances in the fields of computer vision and natural language processing could possibly lead to a higher number of papers, which utilize data from maintenance reports or industrial cameras.
    Currently, we observe only limited research in this area.

    \item Little attention was paid to the interpretability of the proposed solutions and the use of XAI methods.
    The steel industry, which is a critical branch of the economy, as well as a hazardous workplace, requires robust and faithful solutions, so that end-users can trust the prediction of AI models.
    This can be enhanced if the methods used provide explanations of their decisions.

    \item Current research is mainly focused on discovering faults or anomalies in the machinery.
    This approach should decrease the possibility of incoming failures, but does not necessarily improve regular maintenance strategies.
    We argue that more effort should be put into optimizing regular maintenance tasks, so that the share of preventive maintenance actions also decreases.

    \item The transparency and reproducibility of studies should be increased by making the data used for more accessible.
    While it would be preferable to use data from real production facilities, the intellectual property rights possessed by the companies will probably prevent this.
    Similarly, the public availability of the source codes would support researchers and engineers in the implementation and validation of the proposed methods.

    \item More effort from researchers should be put into the implementation and evaluation of the methods from a business perspective.
    The use of AI methods in PdM is a strongly applied field of research; therefore, observable results of the studies should be one of the main priorities.
    This could be achieved by fostering stronger cooperation between researchers and maintenance engineers.

    \item Most researchers work within small, independent teams, lacking substantial collaboration with other scientific groups.
    We believe that cooperation among these research teams has the potential to enhance the overall quality of scientific works.
\end{itemize}

\section{Summary}
\label{sec:summary}
In this survey we conducted a comprehensive analysis of research articles related to the application of AI methods in PdM within the steel industry.
We systematically reviewed and collected 219 research articles related to this field.
The articles were selected from a population of more than 2000 articles, retrieved based on a carefully prepared set of keywords related to AI, PdM, and the steel industry.
The selection process was carried out by independent reviewers, following the PRISMA methodology, to ensure the inclusion of the most relevant studies.
Based on the set of diverse research questions, we in-depthly analyzed the characteristics of all papers, taking into account different dimensions of the research, as well as the characteristics of the data used, along with the business impact of proposed methods.
We believe that this analysis should allow readers to gain insight into different aspects of the subject and be a valuable source of knowledge for both researchers willing to explore the field and practitioners desiring to implement AI methods in their facilities.
In this work we did not use any Generative AI tools. 
All our analyses were conducted manually based on our expertise and with use of automation procedures we programmed. 

Despite the fact that the focus of our survey was on AI-based methods, we also included several papers from related fields such as statistics, signal processing, and analytical models, to provide a more comprehensive view on the subject of the study.
We observed an increasing interest of researchers in the area of AI-based PdM in the steel industry, with a primary focus on the development of deep learning methods.
Research spanned all areas of the steel industry, from sintering to galvanization, but the most attention was devoted to the Hot Rolling Mill and Blast Furnace.
The vast majority of articles used data from sensor readings installed in manufacturing sites, but data in the form of text or images can also be useful in the development of PdM solutions.
Although a large number of studies were conducted, we observed relatively low practical implications in terms of their impact on business.

In summary, this survey confirms that AI-based PdM can bring substantial benefits to the steel industry, including improved operational efficiency, reduced downtime, and improved quality, but the development of intelligent solutions remains a challenging task that requires further studies.
Most importantly, future research should focus on the business aspects of the proposed methods as well as on their practical implications.
In particular, a broader perspective of the subject should be adopted, allowing for the consideration of a specific PdM solution within the maintenance strategy.
Ultimately, the primary goal of PdM is to provide measurable benefits to the efficiency of the steel industry.

\section*{Acknowledgements}
Project XPM is supported by the National Science Centre, Poland (2020/02/Y/ST6/00070), under CHIST-ERA IV programme, which has received funding from the EU Horizon 2020 Research and Innovation Programme, under Grant Agreement no 857925.

\section*{CRediT author statement}

\textbf{Jakub Jakubowski}: Conceptualization, Methodology, Software, Formal analysis, Investigation, Data Curation, Writing - Original Draft, Visualization 
\textbf{Natalia Wojak-Strzelecka}: Conceptualization, Methodology, Formal analysis, Investigation, Data Curation, Writing - Original Draft, Visualization 
\textbf{Rita P. Ribeiro}: Conceptualization, Writing - Original Draft 
\textbf{Sepideh Pashami}: Conceptualization, Writing - Original Draft 
\textbf{Szymon Bobek}: Validation, Writing - Review \& Editing 
\textbf{João Gama}: Conceptualization, Validation, Writing - Review \& Editing 
\textbf{Grzegorz J. Nalepa}: Conceptualization, Methodology, Validation, Writing - Review \& Editing, Supervision, Project administration, Funding acquisition 

 \bibliographystyle{elsarticle-harv} 
 
 \bibliography{references}

\newpage
\appendix
\section{Supplemtary results}
\label{app:details}
In this appendix we provided detailed information regarding survey results.

\subsection{Summary of papers with respect to type of installation}
\label{app:details_installations}

\begin{table}[H]
    \footnotesize
    \centering
    \begin{tabular}{>{\raggedright}p{0.15\columnwidth}>{\raggedright}p{0.17\columnwidth}>{\raggedright\arraybackslash}p{0.6\columnwidth}}
    \toprule
    \textbf{Scope} & \textbf{Asset} & \textbf{Papers} \\
    \midrule
    Installation & & \cite{liu_online_2021} \\
    Subsystem & Sinter Machine & \cite{egorova_diagnostics_2018} \\
    \bottomrule
    \end{tabular}
    \caption{Summary of studies related to the Sinter Plant}
    \label{tab:sinter}
\end{table}

\begin{table}[H]
    \footnotesize
    \centering
    \begin{tabular}{>{\raggedright}p{0.15\columnwidth}>{\raggedright}p{0.17\columnwidth}>{\raggedright\arraybackslash}p{0.6\columnwidth}}
    \toprule
    \textbf{Scope} & \textbf{Asset} & \textbf{Papers} \\
    \midrule
    Entire Plant & & \cite{an_graph-based_2020,chen_hybrid_2023,gao_deep_2022,gao_minimax_2023,han_novel_2022,jiang_research_2019,kacprzyk_new_2011,li_abnormality_2023,li_blast_2023,lian_fault_2010,liu_multi-class_2011,liu_enhanced_2015,liu_structured_2018,lou_fault_2022,lou_structured_2023,lou_data-driven_2023,ouyang_h_and_zeng_j_and_li_y_and_luo_s_fault_2020,pan_fault_2016,pan_robust_2018,saci_autocorrelation_2021,sha_robust_2022,shang_dominant_2017,spirin_expert_2020,tian_novel_2010,wang_anna_and_liu_zuoqian_and_tao_ran_fault_2010,wang_new_2015,wang_blast_2022,xiao_novel_2023,xie_fault_2021,yang_comparison_2015,zhai_multi-block_2020,zhang_knowledge_2012,zhang_fault_2014,zhang_pca-lmnn-based_2016,zhang_tongshuai_and_wang_wei_and_ye_hao_and_huang_dexian_and_zhanghaifeng_and_li_mingliang_fault_2016,zhang_novel_2018,zhang_dynamic_2023,zhao_anomaly_2022,zhou_process_2016,zhou_fault_2020,zhou_data-driven_2021} \\
    Product & Silicon content & \cite{liu_optional_2011,weng_online_2023} \\
    Single Device & Axial fan & \cite{zhang_fault_2021} \\
    & Blowers & \cite{bo_qualitative_2015,fu_digital_2023,raducan_prediction_2020,sun_research_2012} \\
    & Raceway & \cite{puttinger_improving_2019} \\
    & Wall & \cite{xi_remaining_2020,zhang_remaining_2017,zhang_fbm-based_2019,zhang_predicting_2019,zhang_remaining_2021,zhang_intelligent_2022} \\
    Subsystem & Cooling system & \cite{meneganti_fuzzy_1998} \\
    & Gas flow & \cite{zhao_adaptive_2014} \\
    & Hearth & \cite{wang_shuqiao_and_yuan_yan_and_liang_jingjie_and_zhang_zhuofu_design_2022} \\
    & Torpedo car & \cite{chernyi_application_2022,li_research_2020,wang_research_2021,yemelyanov_application_2021} \\
    \bottomrule
    \end{tabular}
    \caption{Summary of studies related to the Blast Furnace}
    \label{tab:bf}
\end{table}

\begin{table}[H]
    \footnotesize
    \centering
    \begin{tabular}{>{\raggedright}p{0.15\columnwidth}>{\raggedright}p{0.17\columnwidth}>{\raggedright\arraybackslash}p{0.6\columnwidth}}
    \toprule
    \textbf{Scope} & \textbf{Asset} & \textbf{Papers} \\
    \midrule
Entire Plant & & \cite{choi_method_2020,wang_approach_2015,wang_detecting_2018} \\
Subsystem & Transformer & \cite{dehghan_marvasti_fault_2014,shiyan_vibration-based_2019} \\
    \bottomrule
    \end{tabular}
    \caption{Summary of studies related to the Electric Arc Furnace}
    \label{tab:eaf}
\end{table}

\begin{table}[H]
    \footnotesize
    \centering
    \begin{tabular}{>{\raggedright}p{0.15\columnwidth}>{\raggedright}p{0.17\columnwidth}>{\raggedright\arraybackslash}p{0.6\columnwidth}}
    \toprule
    \textbf{Scope} & \textbf{Asset} & \textbf{Papers} \\
    \midrule
    Single Device & Ladle & \cite{emelianov_information_2022,liu_anomaly_2021,wang_prediction_2018} \\
    Subsystem & Converter & \cite{chistyakova_intellectual_2019,dong_monitoring_2023} \\
    & Refractory lining & \cite{chistyakova_intellectual_2019} \\
    & Crane & \cite{kim_development_2006} \\
    & Shredder & \cite{neto_deep_2021} \\
    
    \bottomrule
    \end{tabular}
    \caption{Summary of studies related to the Steel Plant}
    \label{tab:steelplant}
\end{table}

\begin{table}[H]
    \footnotesize
    \centering
    \begin{tabular}{>{\raggedright}p{0.15\columnwidth}>{\raggedright}p{0.17\columnwidth}>{\raggedright\arraybackslash}p{0.6\columnwidth}}
    \toprule
    \textbf{Scope} & \textbf{Asset} & \textbf{Papers} \\
    \midrule
    Entire Plant & & \cite{holloway_integration_1991,xu_multi-stage_2023,yang_process_2021,zhou_application_2022} \\
    Single Device & Bearings & \cite{qing_liu_fault_2016,wu_multi-agent_2014} \\
    & Casting roller & \cite{han_framework_2021} \\
    & Crystallizer & \cite{david_j_and_svec_p_and_frischer_r_modelling_2012} \\
    & Exhaust fan & \cite{liu_toward_2023} \\
    & Mold & \cite{hutchison_recognition_2005,lukyanov_intelligent_2015,wu_anomaly_2023} \\
    Subsystem & Casting speed & \cite{zhou_online_2022} \\
    \bottomrule
    \end{tabular}
    \caption{Summary of studies related to the Continuous Casting Machine}
    \label{tab:ccm}
\end{table}

\begin{table}[H]
    \footnotesize
    \centering
    \begin{tabular}{>{\raggedright}p{0.15\columnwidth}>{\raggedright}p{0.17\columnwidth}>{\raggedright\arraybackslash}p{0.6\columnwidth}}
    \toprule
    \textbf{Scope} & \textbf{Asset} & \textbf{Papers} \\
    \midrule
    Entire Plant &  & \cite{han_construction_2022,latham_tool_2023,liu_unsupervised_2023,panagou_explorative_2022,peng_new_2014,wang_fault_2020,weber_learning_2022} \\
    Product & Strip & \cite{ding_high-precision_2023,gerz_comparative_2022,jianliang_research_2020,jing_intelligent_2023,li_self-supervised_2023,mentouri_steel_2020,peng_high-precision_2023,song_application_2022,yang_novel_2023} \\
    Single Device & Backup roll & \cite{yuan_fatigue-damage_2023} \\
    & Bearings & \cite{farina_fault_2015,pan_data-driven_2016,peng_multi-representation_2022} \\
    & Gearbox & \cite{acernese_robust_2021,gao_intelligent_2010,pan_data-driven_2016,peng_multi-representation_2022,russo_fault_2021,sarda_multi-step_2021} \\
    & Looper & \cite{garcia-beltran_causal-based_2004,jiang_optimized_2021,marcu_t_and_koppen-seliger_b_and_stiicher_r_hydraulic_2004,marcu_design_2008} \\
    & Motor & \cite{fagarasan_signal-based_2004,hirata_fault_2015,ma_fault_2023} \\
    & Pump & \cite{goode_plant_2000} \\
    & Work rolls & \cite{anagiannis_energy-based_2020,jakubowski_anomaly_2021,jiao_remaining_2021,kovacic_roll_2019,lebrun_mathematical_2013} \\
    Subsystem & Feedback control system & \cite{li_performance_2020} \\
    & Finishing mill & \cite{ball_model_2020,dong_novel_2022,guo_quality-related_2023,jakubowski_explainable_2021,jiang_recent_2017,jing_correlation_2023,konishi_masami_and_nakano_koichi_application_2010,ma_novel_2021,ma_novel_2021-1,ma_novel_2022,ma_practical_2022,miyabe_methodologies_1988,peng_new_2014,peng_new_kernel_2014,peng_quality-related_2015,peng_quality-based_2016,peng_dynamic_2021,xu_novel_2024,zhang_kpi-based_2017,zhang_extensible_2021,zhang_novel_2022,zhang_comprehensive_2023,zhang_robust_2023,zhang_variational_2023} \\
    & Laminar cooling & \cite{wang_integrated_2023} \\
    & Rolling stand & \cite{acernese_novel_2022,chen_fault_2023,lisounkin_advanced_2002,panagou_feature_2022,ruiz-sarmiento_predictive_2020} \\
    & Roughing mill & \cite{ma_novel_2021-1} \\
    \bottomrule
    \end{tabular}
    \caption{Summary of studies related to the Hot Rolling Mill}
    \label{tab:hrm}
\end{table}

\begin{table}[H]
    \footnotesize
    \centering
    \begin{tabular}{>{\raggedright}p{0.15\columnwidth}>{\raggedright}p{0.17\columnwidth}>{\raggedright\arraybackslash}p{0.6\columnwidth}}
    \toprule
    \textbf{Scope} & \textbf{Asset} & \textbf{Papers} \\
    \midrule
    Product & Strip & \cite{fleischanderl_cnnbased_2022} \\
    Subsystem & Loop control system & \cite{bouhouche_fault_2005,bouhouche_s_and_lahreche_m_and_ziani_s_and_bast_j_quality_2006} \\
    & Waste liquor recovery & \cite{fan_independent_2017} \\
    \bottomrule
    \end{tabular}
    \caption{Summary of studies related to the Pickling Line}
    \label{tab:cpl}
\end{table}

\begin{table}[H]
    \footnotesize
    \centering
    \begin{tabular}{>{\raggedright}p{0.15\columnwidth}>{\raggedright}p{0.17\columnwidth}>{\raggedright\arraybackslash}p{0.6\columnwidth}}
    \toprule
    \textbf{Scope} & \textbf{Asset} & \textbf{Papers} \\
    \midrule
    Entire Plant & & \cite{beden_scro_2021,beden_towards_2023,chen_multi-source_2020,lepenioti_katerina_and_pertselakis_minas_and_bousdekis_alexandros_andlouca_andreas_and_lampathaki_fenareti_and_apostolou_dimitris_andmentzas_gregoris_and_anastasiou_stathis_machine_2020,wang_multi-step-ahead_2023} \\
    Product & Strip & \cite{chen_multi-faceted_2021,yang_sw_and_widodo_a_and_caesarendra_w_and_oh_js_and_shim_mc_and_kim_sj_and_yang_bs_and_lee_wh_support_2009,yang_sliding_2021} \\
    Single Device & Bearings & \cite{ji_fault_2021,shi_intelligent_2022,zhao_fault_2021} \\
    & Motor & \cite{yanbin_sun_and_yi_an_research_2009} \\
    & Work rolls & \cite{fouka_afroditi_and_bousdekis_alexandros_and_lepenioti_katerina_andmentzas_gregoris_real-time_2021,jakubowski_roll_2022,lakshmanan_data_2022,wang_sae-cca-based_2021} \\
    Subsystem & Rolling stand & \cite{lu_prediction_2020,oh_jun-seok_and_kim_hack-eun_case_2015,shi_novel_2022,shin_k-y_and_kwon_w-k_development_2018} \\
    & Tension system & \cite{arinton_neural_2012} \\
    \bottomrule
    \end{tabular}
    \caption{Summary of studies related to the Cold Rolling Mill}
    \label{tab:crm}
\end{table}

\begin{table}[H]
    \footnotesize
    \centering
    \begin{tabular}{>{\raggedright}p{0.15\columnwidth}>{\raggedright}p{0.17\columnwidth}>{\raggedright\arraybackslash}p{0.6\columnwidth}}
    \toprule
    \textbf{Scope} & \textbf{Asset} & \textbf{Papers} \\
    \midrule
    Entire Plant & & \cite{liu_multiblock_2014,liu_quality-relevant_2014,yingwei_zhang_fault_2013} \\
    Product & Strip & \cite{liu_dynamic_2018} \\
    Single Device & Looper & \cite{zhang_fault_2011} \\
    Subsystem & Furnace & \cite{lu_pcasdg_2013,tan_shuai_and_wang_fuli_and_chang_yuqing_and_chen_weidong_and_xujiazhuo_fault_2010,wang_process_2012} \\
    & Furnace temperature system & \cite{lu_pcasdg_2013,wang_process_2012} \\
    & Tension & \cite{qiang_liu_data-based_2011} \\
    \bottomrule
    \end{tabular}
    \caption{Summary of studies related to the Continuous Annealing Line}
    \label{tab:anneal}
\end{table}

\begin{table}[H]
    \footnotesize
    \centering
    \begin{tabular}{>{\raggedright}p{0.15\columnwidth}>{\raggedright}p{0.17\columnwidth}>{\raggedright\arraybackslash}p{0.6\columnwidth}}
    \toprule
    \textbf{Scope} & \textbf{Asset} & \textbf{Papers} \\
    \midrule
    Entire Plant & & \cite{fan_application_2018} \\
    Single Device & Bearings & \cite{chen_application_2021} \\
    & Welder & \cite{bonikila_failure_2022,choi_modeling_2023,meyer_anomaly_2022} \\
    & Zinc bath roller & \cite{simon_health_2021} \\
    Subsystem & Zinc bath & \cite{debon_fault_2012} \\
    \bottomrule
    \end{tabular}
    \caption{Summary of studies related to the Hot Dip Galvanizing Line}
    \label{tab:hdgl}
\end{table}

\subsection{Summary of papers with respect to AI method}

\begin{table}[H]
    \footnotesize
    \centering
    \begin{tabular}{>{\raggedright}p{0.12\columnwidth}>{\raggedright\arraybackslash}p{0.8\columnwidth}}
    \toprule
    \textbf{Method} & \textbf{Papers} \\
    \midrule
    MLP & \cite{arinton_neural_2012,bouhouche_fault_2005,bouhouche_s_and_lahreche_m_and_ziani_s_and_bast_j_quality_2006,chen_multi-source_2020,chen_multi-faceted_2021,chen_application_2021,chen_fault_2023,chernyi_application_2022,chistyakova_intellectual_2019,choi_method_2020,david_j_and_svec_p_and_frischer_r_modelling_2012,dinghui_zhang_researches_2000,egorova_diagnostics_2018,emelianov_information_2022,fleischanderl_cnnbased_2022,gao_deep_2022,gao_minimax_2023,gerz_comparative_2022,han_framework_2021,han_construction_2022,hao_intelligent_2022,jiang_research_2019,jing_intelligent_2023,kacprzyk_new_2011,kim_development_2006,kumar_nonlinear_2022,latham_tool_2023,li_research_2020,li_blast_2023,li_self-supervised_2023,liu_anomaly_2021,liu_toward_2023,lou_fault_2022,lou_data-driven_2023,lu_prediction_2020,ma_fault_2023,marcu_t_and_koppen-seliger_b_and_stiicher_r_hydraulic_2004,marcu_design_2008,meneganti_fuzzy_1998,mentouri_steel_2020,nkonyana_performance_2019,peng_multi-representation_2022,peng_high-precision_2023,perez_visual_2013,qiang_liu_data-based_2011,qing_liu_fault_2016,ray_equipment_1991,shi_intelligent_2022,shi_novel_2022,song_application_2022,wu_multi-agent_2014,xu_novel_2024,yanbin_sun_and_yi_an_research_2009-1,yanbin_sun_and_yi_an_research_2009,yang_sliding_2021,yang_novel_2023,yemelyanov_application_2021,yi_jiangang_and_zeng_peng_analysis_2009,zhao_fault_2021,zhou_application_2022,zhou_online_2022} \\
    CNN & \cite{chernyi_application_2022,fleischanderl_cnnbased_2022,gao_deep_2022,gao_minimax_2023,gerz_comparative_2022,hao_intelligent_2022,jing_intelligent_2023,latham_tool_2023,li_research_2020,li_self-supervised_2023,liu_anomaly_2021,ma_fault_2023,mentouri_steel_2020,peng_multi-representation_2022,peng_high-precision_2023,shi_novel_2022,shi_intelligent_2022,xu_novel_2024,yang_novel_2023,zhao_fault_2021,zhou_application_2022,zhou_online_2022} \\
    LSTM & \cite{chen_multi-source_2020,chen_multi-faceted_2021,choi_method_2020,choi_modeling_2023,fouka_afroditi_and_bousdekis_alexandros_and_lepenioti_katerina_andmentzas_gregoris_real-time_2021,gerz_comparative_2022,han_construction_2022,jakubowski_explainable_2021,jiao_remaining_2021,jing_intelligent_2023,jing_correlation_2023,karagiorgou_unveiling_2019,karagiorgou_making_2020,lakshmanan_data_2022,lepenioti_katerina_and_pertselakis_minas_and_bousdekis_alexandros_andlouca_andreas_and_lampathaki_fenareti_and_apostolou_dimitris_andmentzas_gregoris_and_anastasiou_stathis_machine_2020,ma_practical_2022,wu_anomaly_2023,zhang_intelligent_2022,zhou_application_2022,zhou_online_2022} \\
    AE & \cite{chen_hybrid_2023,choi_modeling_2023,jakubowski_explainable_2021,jakubowski_anomaly_2021,jakubowski_roll_2022,jing_intelligent_2023,li_self-supervised_2023,li_abnormality_2023,liu_toward_2023,meyer_anomaly_2022,wang_sae-cca-based_2021,wu_anomaly_2023,yang_novel_2023,zhang_tongshuai_and_wang_wei_and_ye_hao_and_huang_dexian_and_zhanghaifeng_and_li_mingliang_fault_2016,zhang_robust_2023,zhou_application_2022} \\
    GRU & \cite{chen_multi-source_2020,chen_multi-faceted_2021,li_abnormality_2023,ma_practical_2022,ma_fault_2023,ouyang_h_and_zeng_j_and_li_y_and_luo_s_fault_2020,wang_research_2021,wu_anomaly_2023,xu_multi-stage_2023,xu_novel_2024,zhang_intelligent_2022,zhang_comprehensive_2023} \\
    RNN & \cite{chen_multi-source_2020,chen_multi-faceted_2021,choi_method_2020,marcu_design_2008,yang_sliding_2021} \\
    GAN & \cite{peng_high-precision_2023,xie_fault_2021,zhang_robust_2023,zhao_fault_2021} \\
    Attention & \cite{hao_intelligent_2022,shi_intelligent_2022,wang_research_2021} \\
    DBN & \cite{ji_fault_2021,jianliang_research_2020} \\
    DDQN & \cite{acernese_novel_2022,neto_deep_2021} \\
    Transformer & \cite{han_construction_2022,ma_fault_2023} \\
    RBM & \cite{dong_monitoring_2023,liu_unsupervised_2023} \\
    RBF NN & \cite{perez_visual_2013} \\
    BLS & \cite{zhang_extensible_2021} \\
    GNN & \cite{han_construction_2022} \\
    
    \bottomrule
    \end{tabular}
    \caption{Summary of studies utilizing Neural Network methods}
    \label{tab:neural}
\end{table}

\begin{table}[H]
    \footnotesize
    \centering
    \begin{tabular}{>{\raggedright}p{0.2\columnwidth}>{\raggedright\arraybackslash}p{0.75\columnwidth}}
    \toprule
    \textbf{Method} & \textbf{Papers} \\
    \midrule
    Gradient Boosting & \cite{ding_high-precision_2023,liu_online_2021,lu_prediction_2020,panagou_feature_2022,tiensuu_intelligent_2020,wang_fault_2020,wang_multi-step-ahead_2023,xie_fault_2021,xu_novel_2024} \\
    RF & \cite{chen_multi-source_2020,chen_application_2021,hutchison_approach_2014,nkonyana_performance_2019,wang_multi-step-ahead_2023} \\
    CART & \cite{debon_fault_2012,song_application_2022} \\
    Decision Tree & \cite{hutchison_approach_2014} \\
    Isolation Forest & \cite{gerz_comparative_2022} \\
    
    \bottomrule
    \end{tabular}
    \caption{Summary of studies utilizing Tree-based methods}
    \label{tab:tree}
\end{table}

\begin{table}[H]
    \footnotesize
    \centering
    \begin{tabular}{>{\raggedright}p{0.15\columnwidth}>{\raggedright\arraybackslash}p{0.8\columnwidth}}
    \toprule
    \textbf{Method} & \textbf{Papers} \\
    \midrule
    SVM & \cite{chen_multi-source_2020,fu_digital_2023,gerz_comparative_2022,hutchison_approach_2014,jiang_optimized_2021,konishi_masami_and_nakano_koichi_application_2010,li_blast_2023,liu_multi-class_2011,liu_optional_2011,lou_fault_2022,mentouri_steel_2020,nkonyana_performance_2019,ouyang_h_and_zeng_j_and_li_y_and_luo_s_fault_2020,russo_fault_2021,tian_novel_2010,wang_anna_and_liu_zuoqian_and_tao_ran_fault_2010,wang_new_2015,wang_detecting_2018,wang_prediction_2018,wang_blast_2022,weber_learning_2022,yang_sw_and_widodo_a_and_caesarendra_w_and_oh_js_and_shim_mc_and_kim_sj_and_yang_bs_and_lee_wh_support_2009,yang_comparison_2015,zhai_multi-block_2020,zhang_novel_2022} \\
    OCSVM & \cite{gerz_comparative_2022,jiang_optimized_2021,ouyang_h_and_zeng_j_and_li_y_and_luo_s_fault_2020,russo_fault_2021,wang_detecting_2018,wang_prediction_2018,wang_blast_2022,weber_learning_2022,zhai_multi-block_2020,zhang_novel_2022} \\
    SVR & \cite{liu_enhanced_2015,lu_prediction_2020,song_application_2022,wang_multi-step-ahead_2023,zhang_fault_2021} \\

    \bottomrule
    \end{tabular}
    \caption{Summary of studies utilizing Support Vectors methods}
    \label{tab:suport_vectors}
\end{table}

\begin{table}[H]
    \footnotesize
    \centering
    \begin{tabular}{>{\raggedright}p{0.3\columnwidth}>{\raggedright\arraybackslash}p{0.65\columnwidth}}
    \toprule
    \textbf{Method} & \textbf{Papers} \\
    \midrule
    Gaussian Mixture Model & \cite{chen_hybrid_2023,han_novel_2022,peng_quality-related_2015,peng_dynamic_2021,xiao_novel_2023,zhang_dynamic_2023} \\
    Hidden Markov Model & \cite{sarda_multi-step_2021,simon_health_2021,wang_approach_2015} \\
    Causal Graph Model & \cite{garcia-beltran_causal-based_2004,yang_process_2021} \\
    Bayes & \cite{fouka_afroditi_and_bousdekis_alexandros_and_lepenioti_katerina_andmentzas_gregoris_real-time_2021,lian_fault_2010,ma_novel_2021-1,peng_quality-based_2016,ruiz-sarmiento_predictive_2020,zhang_variational_2023} \\
    Bayesian Network & \cite{lian_fault_2010} \\
    Gaussian Process & \cite{wang_prediction_2018} \\
    Discrete Bayes Filter & \cite{ruiz-sarmiento_predictive_2020} \\
    Bayesian Changepoint Detection & \cite{fouka_afroditi_and_bousdekis_alexandros_and_lepenioti_katerina_andmentzas_gregoris_real-time_2021} \\
    Bayesian State Space Model & \cite{zhang_variational_2023} \\
    \bottomrule
    \end{tabular}
    \caption{Summary of studies utilizing Probabilistic methods}
    \label{tab:probabilistic}
\end{table}

\begin{table}[H]
    \footnotesize
    \centering
    \begin{tabular}{>{\raggedright}p{0.3\columnwidth}>{\raggedright\arraybackslash}p{0.65\columnwidth}}
    \toprule
    \textbf{Method} & \textbf{Papers} \\
    \midrule
    k-Means & \cite{gerz_comparative_2022,shin_k-y_and_kwon_w-k_development_2018,wang_detecting_2018} \\
    DBSCAN & \cite{gerz_comparative_2022,jin_varying-scale_2023} \\
    Qualitative Trend Clustering & \cite{bo_qualitative_2015} \\
    mean-shifted clustering & \cite{shin_k-y_and_kwon_w-k_development_2018} \\
    density-based clustering & \cite{shin_k-y_and_kwon_w-k_development_2018} \\
    \bottomrule
    \end{tabular}
    \caption{Summary of studies utilizing Clustering methods}
    \label{tab:cluster}
\end{table}

\begin{table}[H]
    \centering
    \footnotesize
     \begin{tabular}{>{\raggedright}p{0.2\columnwidth}|>{\raggedright\arraybackslash}p{0.75\columnwidth}}
    \toprule
    \textbf{Method} & \textbf{Papers} \\
    \midrule
    Nearest Neighbors & \cite{liu_optional_2011}, \cite{zhao_adaptive_2014}, \cite{zhang_pca-lmnn-based_2016}, \cite{mentouri_steel_2020}, \cite{dong_novel_2022}, \cite{fu_digital_2023}, \cite{ma_fault_2023}, \cite{wang_multi-step-ahead_2023} \\
    Signal Processing & \cite{fagarasan_signal-based_2004}, \cite{sun_research_2012}, \cite{dehghan_marvasti_fault_2014}, \cite{pan_data-driven_2016}, \cite{shiyan_vibration-based_2019} \\
    Digital Twin & \cite{han_framework_2021}, \cite{panagou_explorative_2022}, \cite{panagou_feature_2022}, \cite{fu_digital_2023} \\
    Linear & \cite{debon_fault_2012}, \cite{shang_dominant_2017}, \cite{fu_digital_2023} \\
    MTL & \cite{ma_novel_2021}, \cite{ma_novel_2022} \\
    \bottomrule
    \end{tabular}
    \caption{Summary of studies utilizing miscellaneous ML methods}
    \label{tab:other_methods}
\end{table}

\begin{table}[ht]
   \centering
   \footnotesize
    \begin{tabular}{>{\raggedright}p{0.2\columnwidth}|>{\raggedright\arraybackslash}p{0.70\columnwidth}}
   \toprule
   \textbf{Method} & \textbf{Papers} \\
   \midrule
   Fuzzy &  \cite{ganeha_b__k_n_umesh_condition_2019}, \cite{hutchison_recognition_2005}, \cite{ganeha_b__k_n_umesh_condition_2019}, \cite{lisounkin_advanced_2002}, \\
   Rules & \cite{gao_intelligent_2010}, \cite{miyabe_methodologies_1988}, \cite{ray_artificial_1989}, \cite{slany_consistency_1996} \\
   Expert System & \cite{chistyakova_intellectual_2019}, \cite{dinghui_zhang_researches_2000}, \cite{ray_artificial_1989}, \cite{ray_equipment_1991},  \cite{spirin_expert_2020} \\
   Ontology & \cite{beden_scro_2021},  \cite{beden_towards_2023}, \cite{cao_core_2022}, \cite{zhang_knowledge_2012} \\
   \bottomrule
   \end{tabular}
   \caption{Summary of studies utilizing knowledge-based methods}
   \label{tab:knowledge-based}
\end{table}

\begin{table}[H]
    \footnotesize
    \centering
    \begin{tabular}{>{\raggedright}p{0.25\columnwidth}>{\raggedright\arraybackslash}p{0.7\columnwidth}}
    \toprule
    \textbf{Method} & \textbf{Papers} \\
    \midrule
    ICA & \cite{fan_independent_2017,fan_application_2018,peng_new_2014,peng_new_2014,tan_shuai_and_wang_fuli_and_chang_yuqing_and_chen_weidong_and_xujiazhuo_fault_2010,wang_process_2012,yingwei_zhang_fault_2013,zhou_data-driven_2021} \\
    FBM & \cite{xi_remaining_2020,zhang_remaining_2017,zhang_fbm-based_2019,zhang_predicting_2019,zhang_remaining_2021} \\
    CCA & \cite{liu_dynamic_2018,peng_new_2014,wang_sae-cca-based_2021} \\
    CVA & \cite{chen_fault_2023,lou_structured_2023,wang_blast_2022} \\
    PCP & \cite{pan_fault_2016,pan_robust_2018} \\
    Mahalanobis Distance & \cite{an_graph-based_2020,shin_k-y_and_kwon_w-k_development_2018} \\
    Slow Features & \cite{dong_novel_2022,zhai_multi-block_2020} \\
    Scheffee Test & \cite{lisounkin_advanced_2002} \\
    Markov Chain & \cite{zhang_fbm-based_2019} \\
    MCD & \cite{acernese_robust_2021} \\
    Grey Model & \cite{zhao_anomaly_2022} \\
    Mann-Kendall & \cite{weng_online_2023} \\
    \bottomrule
    \end{tabular}
    \caption{Summary of studies utilizing Statistical methods}
    \label{tab:statistics}
\end{table}

\begin{table}[H]
    \footnotesize
    \centering
    \begin{tabular}{>{\raggedright}p{0.25\columnwidth}>{\raggedright\arraybackslash}p{0.7\columnwidth}}
    \toprule
    \textbf{Method} & \textbf{Papers} \\
    \midrule
PCA and AI & \cite{yang_sw_and_widodo_a_and_caesarendra_w_and_oh_js_and_shim_mc_and_kim_sj_and_yang_bs_and_lee_wh_support_2009}, \cite{lu_pcasdg_2013}, \cite{zhang_pca-lmnn-based_2016}, \cite{shang_dominant_2017}, \cite{egorova_diagnostics_2018}, \cite{lu_prediction_2020}, \cite{jiang_optimized_2021}, \cite{ma_novel_2021-1}, \cite{peng_dynamic_2021}, \cite{han_novel_2022}, \cite{kumar_nonlinear_2022}, \cite{meyer_anomaly_2022}, \cite{weber_learning_2022}, \cite{xiao_novel_2023}, \cite{xu_novel_2024} \\
PCA & \cite{zhang_fault_2011}, \cite{shigemori_hiroyasu_quality_2013}, \cite{liu_multiblock_2014}, \cite{zhang_fault_2014}, \cite{hirata_fault_2015}, \cite{zhou_process_2016}, \cite{liu_structured_2018} \\
PCA and Statistics & \cite{wang_process_2012}, \cite{yingwei_zhang_fault_2013}, \cite{peng_new_2014}, \cite{peng_new_2014}, \cite{zhou_data-driven_2021}, \cite{zhao_anomaly_2022} \\
PLS & \cite{liu_quality-relevant_2014}, \cite{jiang_recent_2017}, \cite{deepak_multivariate_2020}, \cite{zhou_fault_2020}, \cite{straat_industry_2022}, \cite{guo_quality-related_2023} \\
PLS and AI & \cite{peng_quality-related_2015}, \cite{peng_quality-based_2016}, \cite{chen_application_2021}, \cite{zhang_extensible_2021} \\
t-SNE & \cite{perez_visual_2013} \\
NPE & \cite{sha_robust_2022} \\
    \bottomrule
    \end{tabular}
    \caption{Summary of studies utilizing dimensionality reduction methods}
    \label{tab:dim_reduc}
\end{table}

\begin{table}[H]
    \footnotesize
    \centering
    \begin{tabular}{>{\raggedright}p{0.15\columnwidth}>{\raggedright}p{0.15\columnwidth}>{\raggedright\arraybackslash}p{0.65\columnwidth}}
    \toprule
    \textbf{Category} & \textbf{Method} & \textbf{Papers} \\
    \midrule
    Optimization & GA & \cite{ji_fault_2021,jing_intelligent_2023,qing_liu_fault_2016,yang_comparison_2015} \\
     & PSO & \cite{wang_shuqiao_and_yuan_yan_and_liang_jingjie_and_zhang_zhuofu_design_2022,yang_comparison_2015} \\
     & Genetic Programming & \cite{kovacic_roll_2019} \\
     & CSSA & \cite{zhang_fault_2021} \\
     & Baysesian Optimization & \cite{jakubowski_anomaly_2021,jakubowski_explainable_2021} \\

    Analytical Model &        & \cite{lisounkin_advanced_2002}, \cite{lebrun_mathematical_2013}, \cite{lukyanov_intelligent_2015}, \cite{anagiannis_energy-based_2020}, \cite{ball_model_2020}, \cite{emelianov_information_2022}, \cite{jakubowski_roll_2022}, \cite{wang_shuqiao_and_yuan_yan_and_liang_jingjie_and_zhang_zhuofu_design_2022}, \cite{wang_integrated_2023}, \cite{yuan_fatigue-damage_2023} \\

    Signal Processing &        & \cite{fagarasan_signal-based_2004}, \cite{sun_research_2012}, \cite{dehghan_marvasti_fault_2014}, \cite{pan_data-driven_2016}, \cite{shiyan_vibration-based_2019} \\

    \bottomrule
    \end{tabular}
    \caption{Summary of studies utilizing other techniques}
    \label{tab:optimization}
\end{table}

\begin{table}[H]
    \footnotesize
    \centering
    \begin{tabular}{>{\raggedright}p{0.2\columnwidth}>{\raggedright\arraybackslash}p{0.75\columnwidth}}
    \toprule
    \textbf{Method} & \textbf{Papers} \\
    \midrule
Contribution Plot & \cite{guo_quality-related_2023}, \cite{han_novel_2022}, \cite{kumar_nonlinear_2022}, \cite{liu_structured_2018}, \cite{zhang_fault_2011}, \cite{zhao_anomaly_2022} \\
  Counterfactuals &  \cite{jakubowski_roll_2022} \\
             SHAP &  \cite{ding_high-precision_2023}, \cite{jakubowski_anomaly_2021}, \cite{jakubowski_explainable_2021}, \cite{tiensuu_intelligent_2020} \\
    \bottomrule
    \end{tabular}
    \caption{Summary of studies utilizing XAI methods}
    \label{tab:xa}
\end{table} 

    

\subsection{Predictive Maintenance tasks}
\begin{table}[H]
    \footnotesize
    \centering
    \begin{tabular}{>{\raggedright}p{0.14\columnwidth}>{\raggedright\arraybackslash}p{0.83\columnwidth}}
    \toprule
     \textbf{Task} &    \textbf{Papers} \\
        \midrule
        Anomaly Detection & \cite{acernese_robust_2021,choi_modeling_2023,gerz_comparative_2022,jakubowski_explainable_2021,jakubowski_anomaly_2021,jiang_research_2019,jin_varying-scale_2023,junichi_t_and_yukinori_i_condition_2020,konishi_masami_and_nakano_koichi_application_2010,li_self-supervised_2023,li_abnormality_2023,liu_anomaly_2021,liu_online_2021,liu_unsupervised_2023,meyer_anomaly_2022,pan_fault_2016,saci_autocorrelation_2021,sarda_multi-step_2021,spirin_expert_2020,wang_approach_2015,wang_detecting_2018,wang_prediction_2018,wang_integrated_2023,weber_learning_2022,weng_online_2023,wu_anomaly_2021,wu_anomaly_2023,xiao_novel_2023,yingwei_zhang_fault_2013,zhao_adaptive_2014,zhao_anomaly_2022,zhou_process_2016,zhou_application_2022,zhou_online_2022} \\
 Condition Monitoring &   \cite{oh_jun-seok_and_kim_hack-eun_case_2015,shiyan_vibration-based_2019,wang_shuqiao_and_yuan_yan_and_liang_jingjie_and_zhang_zhuofu_design_2022,yemelyanov_application_2021} \\
     Defect Detection & \cite{fleischanderl_cnnbased_2022,hao_intelligent_2022,hutchison_approach_2014,yang_novel_2023} \\
      Fault Detection & \cite{acernese_novel_2022,an_graph-based_2020,arinton_neural_2012,bonikila_failure_2022,bouhouche_fault_2005,bouhouche_s_and_lahreche_m_and_ziani_s_and_bast_j_quality_2006,chen_multi-source_2020,chen_multi-faceted_2021,chen_fault_2023,debon_fault_2012,deepak_multivariate_2020,dehghan_marvasti_fault_2014,ding_high-precision_2023,dong_novel_2022,fagarasan_signal-based_2004,fan_independent_2017,fan_application_2018,farina_fault_2015,fu_digital_2023,ganeha_b__k_n_umesh_condition_2019,guo_quality-related_2023,hirata_fault_2015,holloway_integration_1991,hutchison_recognition_2005,jakubowski_roll_2022,ji_fault_2021,jiang_recent_2017,jiang_optimized_2021,li_performance_2020,li_research_2020,li_blast_2023,lisounkin_advanced_2002,liu_quality-relevant_2014,liu_structured_2018,lou_structured_2023,lu_prediction_2020,lukyanov_intelligent_2015,ma_novel_2021,ma_novel_2021-1,ma_novel_2022,marcu_t_and_koppen-seliger_b_and_stiicher_r_hydraulic_2004,marcu_design_2008,meneganti_fuzzy_1998,ouyang_h_and_zeng_j_and_li_y_and_luo_s_fault_2020,pan_robust_2018,panagou_feature_2022,peng_new_2014,peng_quality-related_2015,peng_quality-based_2016,peng_dynamic_2021,peng_high-precision_2023,perez_visual_2013,puttinger_improving_2019,qiang_liu_data-based_2011,ray_equipment_1991,russo_fault_2021,sha_robust_2022,shigemori_hiroyasu_quality_2013,straat_industry_2022,tan_shuai_and_wang_fuli_and_chang_yuqing_and_chen_weidong_and_xujiazhuo_fault_2010,wang_process_2012,wang_fault_2020,wang_sae-cca-based_2021,wang_blast_2022,wang_multi-step-ahead_2023,xie_fault_2021,xu_multi-stage_2023,yang_sw_and_widodo_a_and_caesarendra_w_and_oh_js_and_shim_mc_and_kim_sj_and_yang_bs_and_lee_wh_support_2009,yang_sliding_2021,yang_process_2021,zhai_multi-block_2020,zhang_fault_2014,zhang_tongshuai_and_wang_wei_and_ye_hao_and_huang_dexian_and_zhanghaifeng_and_li_mingliang_fault_2016,zhang_kpi-based_2017,zhang_fault_2021,zhang_novel_2022,zhang_variational_2023,zhang_dynamic_2023,zhang_robust_2023} \\
      Fault Diagnosis & \cite{bo_qualitative_2015,chernyi_application_2022,chistyakova_intellectual_2019,deepak_multivariate_2020,dinghui_zhang_researches_2000,dong_monitoring_2023,egorova_diagnostics_2018,gao_intelligent_2010,gao_deep_2022,emelianov_information_2022,gao_minimax_2023,garcia-beltran_causal-based_2004,han_construction_2022,han_novel_2022,jianliang_research_2020,jing_intelligent_2023,jing_correlation_2023,kumar_nonlinear_2022,lian_fault_2010,lisounkin_advanced_2002,liu_optional_2011,liu_multi-class_2011,liu_multiblock_2014,liu_dynamic_2018,lou_fault_2022,lou_data-driven_2023,lu_pcasdg_2013,ma_fault_2023,mentouri_steel_2020,miyabe_methodologies_1988,nkonyana_performance_2019,pan_data-driven_2016,peng_multi-representation_2022,ray_artificial_1989,shang_dominant_2017,shi_novel_2022,shi_intelligent_2022,sun_research_2012,tian_novel_2010,wang_anna_and_liu_zuoqian_and_tao_ran_fault_2010,wang_new_2015,wang_research_2021,xu_novel_2024,yanbin_sun_and_yi_an_research_2009,yanbin_sun_and_yi_an_research_2009-1,yang_comparison_2015,yi_jiangang_and_zeng_peng_analysis_2009,zhang_fault_2011,zhang_pca-lmnn-based_2016,zhao_fault_2021} \\
      Fault Isolation &   \cite{fagarasan_signal-based_2004,zhang_extensible_2021} \\
   Process Monitoring & \cite{chen_hybrid_2023,liu_structured_2018,peng_new_2014,peng_new_2014,zhang_novel_2018,zhang_comprehensive_2023} \\
  Root Cause Analysis & \cite{han_novel_2022,kumar_nonlinear_2022,latham_tool_2023,ma_practical_2022,tiensuu_intelligent_2020} \\
 Fault Identification &  \cite{liu_toward_2023,zhang_comprehensive_2023,zhou_fault_2020,zhou_data-driven_2021} \\
    Health Estimation & \cite{chen_application_2021,david_j_and_svec_p_and_frischer_r_modelling_2012,fouka_afroditi_and_bousdekis_alexandros_and_lepenioti_katerina_andmentzas_gregoris_real-time_2021,lebrun_mathematical_2013,qing_liu_fault_2016,ruiz-sarmiento_predictive_2020,wu_multi-agent_2014,yuan_fatigue-damage_2023} \\
   Quality Prediction &  \cite{liu_enhanced_2015,song_application_2022,straat_industry_2022} \\
Remaining Useful Life &    \cite{anagiannis_energy-based_2020,ball_model_2020,choi_method_2020,goode_plant_2000,han_framework_2021,jiao_remaining_2021,karagiorgou_making_2020,kim_development_2006,kovacic_roll_2019,lakshmanan_data_2022,lepenioti_katerina_and_pertselakis_minas_and_bousdekis_alexandros_andlouca_andreas_and_lampathaki_fenareti_and_apostolou_dimitris_andmentzas_gregoris_and_anastasiou_stathis_machine_2020,qing_liu_fault_2016,raducan_prediction_2020,simon_health_2021,wu_multi-agent_2014,xi_remaining_2020,zhang_remaining_2017,zhang_fbm-based_2019,zhang_predicting_2019,zhang_remaining_2021,zhang_intelligent_2022} \\
           Scheduling &   \cite{holloway_integration_1991,neto_deep_2021,slany_consistency_1996} \\

    \bottomrule
    \end{tabular}
    \caption{Summary of studies with respect to PdM task considered}
    \label{tab:pdm_tasks}
\end{table}

\subsection{Summary of papers concerning the bussiness impact of research}
\begin{table}[H]
    \footnotesize
    \centering
    \begin{tabular}{>{\raggedright}p{0.2\columnwidth}>{\raggedright\arraybackslash}p{0.75\columnwidth}}
\toprule
\textbf{Criterion} & \textbf{Papers} \\
\midrule
Collaboration & \cite{acernese_robust_2021,beden_scro_2021,beden_towards_2023,bonikila_failure_2022,bouhouche_fault_2005,chen_application_2021,chernyi_application_2022,chistyakova_intellectual_2019,choi_modeling_2023,david_j_and_svec_p_and_frischer_r_modelling_2012,dehghan_marvasti_fault_2014,emelianov_information_2022,fagarasan_signal-based_2004,fan_independent_2017,fu_digital_2023,gao_intelligent_2010,garcia-beltran_causal-based_2004,goode_plant_2000,holloway_integration_1991,hutchison_recognition_2005,jakubowski_explainable_2021,jakubowski_anomaly_2021,jakubowski_roll_2022,jin_varying-scale_2023,kacprzyk_new_2011,kovacic_roll_2019,latham_tool_2023,lepenioti_katerina_and_pertselakis_minas_and_bousdekis_alexandros_andlouca_andreas_and_lampathaki_fenareti_and_apostolou_dimitris_andmentzas_gregoris_and_anastasiou_stathis_machine_2020,li_performance_2020,lisounkin_advanced_2002,lou_fault_2022,lukyanov_intelligent_2015,marcu_t_and_koppen-seliger_b_and_stiicher_r_hydraulic_2004,miyabe_methodologies_1988,oh_jun-seok_and_kim_hack-eun_case_2015,ouyang_h_and_zeng_j_and_li_y_and_luo_s_fault_2020,panagou_feature_2022,puttinger_improving_2019,raducan_prediction_2020,ray_artificial_1989,russo_fault_2021,sarda_multi-step_2021,shigemori_hiroyasu_quality_2013,shiyan_vibration-based_2019,simon_health_2021,tiensuu_intelligent_2020,wang_fault_2020,wang_shuqiao_and_yuan_yan_and_liang_jingjie_and_zhang_zhuofu_design_2022,weber_learning_2022,yang_process_2021,yemelyanov_application_2021,yuan_fatigue-damage_2023,zhang_fault_2014,zhang_tongshuai_and_wang_wei_and_ye_hao_and_huang_dexian_and_zhanghaifeng_and_li_mingliang_fault_2016,zhang_comprehensive_2023,zhou_fault_2020} \\
Implemented & \cite{bouhouche_fault_2005,chen_application_2021,fagarasan_signal-based_2004,garcia-beltran_causal-based_2004,holloway_integration_1991,hutchison_recognition_2005,jin_varying-scale_2023,kim_development_2006,lisounkin_advanced_2002,lukyanov_intelligent_2015,panagou_explorative_2022,peng_dynamic_2021,raducan_prediction_2020,shigemori_hiroyasu_quality_2013,shin_k-y_and_kwon_w-k_development_2018,tiensuu_intelligent_2020,wang_shuqiao_and_yuan_yan_and_liang_jingjie_and_zhang_zhuofu_design_2022,yemelyanov_application_2021} \\
Benefits presented & \cite{choi_modeling_2023,han_framework_2021,lukyanov_intelligent_2015} \\
\bottomrule
\end{tabular}
    \caption{Summary of papers concerning the business impact of research}
    \label{tab:rq5}
\end{table}








\end{document}